\documentclass[sigconf,nonacm]{acmart}

\AtBeginDocument{%
  }

\setcopyright{acmlicensed}
\copyrightyear{2026}
\acmYear{2026}
\acmDOI{XXXXXXX.XXXXXXX}

\acmConference[MM '26]{ACM Multimedia}{November 10--14, 2026}{Rio de Janeiro, Brazil}

\renewcommand\footnotetextcopyrightpermission[1]{} 
\author{Jian Yu}
\affiliation{%
  \institution{Nanjing University of Science and Technology}
  \country{}
}

\author{Fei Shen}
\authornote{Project Lead}
\affiliation{%
  \institution{National University of Singapore}
  \country{}
}

\author{Cong Wang}
\affiliation{%
  \institution{Nanjing University}
  \country{}
}

\author{Yi Xin}
\affiliation{%
  \institution{Nanjing University}
  \country{}
}

\author{Si Shen}
\affiliation{%
  \institution{Nanjing University of Science and Technology}
  \country{}
}

\author{Xiaoyu Du}
\affiliation{%
  \institution{Nanjing University of Science and Technology}
  \country{}
}

\author{Jinhui Tang}
\affiliation{%
  \institution{Nanjing Forestry University}
  \country{}
}

\usepackage{booktabs} 
\usepackage{appendix}
\usepackage{booktabs}
\usepackage{multirow}
\usepackage{amsmath} 

\usepackage[table]{xcolor} 
\usepackage[most]{tcolorbox} 

\usepackage{enumitem}

\usepackage[ruled]{algorithm2e} 

\SetAlFnt{\small}
\SetAlCapFnt{\small}
\SetAlCapNameFnt{\small}
\SetAlCapHSkip{0pt}

\usepackage{pifont}  
\usepackage{graphicx} 
\usepackage{xcolor}
\newcommand{\cmark}{\ding{51}} 
\newcommand{\xmark}{\textcolor{gray}{\ding{55}}}

\begin{document}

\title{VersaVogue: Visual Expert Orchestration and Preference Alignment for Unified Fashion Synthesis}

\begin{abstract}

Diffusion models have driven remarkable advancements in fashion image generation, yet prior works usually treat garment generation and virtual dressing as separate problems, limiting their flexibility in real-world fashion workflows. Moreover, fashion image synthesis under multi-source heterogeneous conditions remains challenging, as existing methods typically rely on simple feature concatenation or static layer-wise injection, which often causes attribute entanglement and semantic interference.
To address these issues, we propose VersaVogue, a unified framework for multi-condition controllable fashion synthesis that jointly supports garment generation and virtual dressing, corresponding to the design and showcase stages of the fashion lifecycle. Specifically, we introduce a trait-routing attention (TA) module that leverages a mixture-of-experts mechanism to dynamically route condition features to the most compatible experts and generative layers, enabling disentangled injection of visual attributes such as texture, shape, and color. 
To further improve realism and controllability, we develop an automated multi-perspective preference optimization (MPO) pipeline that constructs preference data without human annotation or task-specific reward models. By combining evaluators of content fidelity, textual alignment, and perceptual quality, MPO identifies reliable preference pairs, which are then used to optimize the model via direct preference optimization (DPO). 
Extensive experiments on both garment generation and virtual dressing benchmarks demonstrate that VersaVogue consistently outperforms existing methods in visual fidelity, semantic consistency, and fine-grained controllability.
\end{abstract}



\maketitle

\begin{figure}
    \centering
    \includegraphics[width=0.85\linewidth]{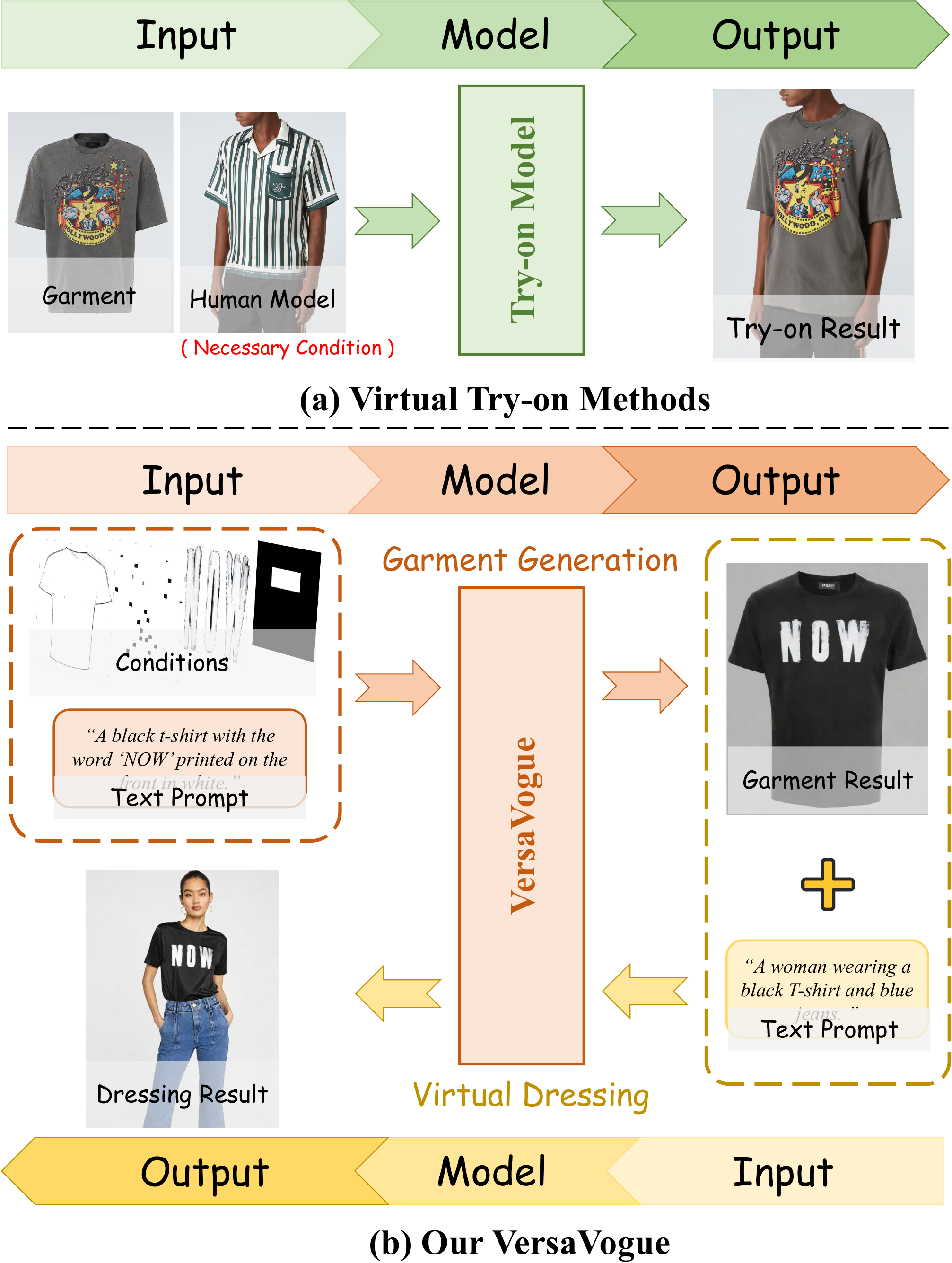}
    \caption{Differences in workflows and input conditions between virtual try-on methods and our VersaVogue. (a) Try-on methods are restricted to the showcase stage and strictly rely on an explicit model image. (b) VersaVogue establishes a unified workflow from garment design to virtual dressing, synthesizing on-body results directly from text descriptions.}
    \vspace{-0.5cm} 
    \label{fig:intro}
    
\end{figure}

\section{Introduction}
Recent advances in generative models~\cite{podell2023sdxl,rombach2022high,peebles2023scalable} have significantly expanded the potential of high-fidelity image synthesis for fashion applications. Among them, garment generation aims to synthesize detailed apparel designs from multimodal conditions, while virtual dressing focuses on rendering photorealistic on-body showcases. These two tasks correspond to the design and showcase stages of the fashion lifecycle, respectively. A unified framework that bridges them would not only streamline traditional fashion workflows by reducing manual effort and production cost, but also improve digital retail through more flexible and personalized visual content. 

Despite their close connection, existing studies address garment generation and virtual dressing as separate problems using fragmented, task-specific pipelines~\cite{zhu2024logosticker,kim2024stableviton,xu2025ootdiffusion}, as illustrated in Figure~\ref{fig:intro} (a). For example, text-driven garment generation methods such as DiffCloth~\cite{zhang2023diffcloth} rely on structural priors to anchor semantic attributes, while virtual dressing methods such as IMAGDressing~\cite{shen2025imagdressing} and StableGarment~\cite{wang2024stablegarment} design specialized modules, \emph{e.g.}, parallel UNets or hybrid attention, to inject garment references. Although effective for their individual settings, these approaches lack a unified mechanism for handling both tasks under a common formulation.

A key obstacle to such unification lies in the need to process complex multi-source heterogeneous conditions while preserving fine-grained controllability. Since both garment generation and virtual dressing fundamentally require accurate attribute preservation under diverse visual and textual constraints, we reformulate them as a unified multi-condition controllable generation problem. However, existing general-purpose conditioning strategies still suffer from inherent limitations. Some methods~\cite{ye2023ip,mou2024t2i} employ lightweight adapters with sequential stacking, which often capture only superficial semantics and fail to preserve high-fidelity consistency with the input conditions. Other approaches based on static layer-wise injection or direct feature concatenation~\cite{lin2025dreamfit,zhao2023uni} lack the flexibility to coordinate heterogeneous conditions, frequently leading to attribute entanglement, semantic interference, and unsatisfactory synthesis quality. Even more specialized solutions such as IMAGGarment~\cite{shen2025imaggarment} rely on multi-stage pipelines to decouple attributes, sacrificing efficiency and end-to-end coherence.

To address these challenges, we propose \textbf{VersaVogue}, a unified framework for fashion synthesis under complex multi-source heterogeneous conditions from Figure~\ref{fig:intro} (b). 
The core idea is to replace static condition injection with dynamic expert orchestration and further improve generation quality through automated preference alignment. Specifically, we introduce a trait-routing attention (TA) module built upon a mixture-of-experts (MoE) mechanism. Given multiple condition inputs, TA dynamically routes attribute-aware features to the most compatible layers and experts in the generative model, enabling more precise disentanglement and adaptive injection of visual traits such as texture, shape, and color. In this way, VersaVogue mitigates feature conflicts among heterogeneous inputs while maintaining a unified end-to-end inference pipeline.

In addition, high controllability alone does not guarantee perceptual realism or alignment with human preference. Recent direct preference optimization (DPO) methods~\cite{rafailov2023direct,wallace2024diffusion} provide an effective way to refine generation behavior, but they typically depend on human-annotated preference data, which is costly to collect and inevitably subjective. We therefore develop a fully automated multi-perspective preference optimization (MPO) pipeline. By jointly evaluating candidate samples from the perspectives of content fidelity, textual alignment, and aesthetic quality, MPO constructs reliable preference pairs without human annotation or task-specific reward models. Based on these automatically curated preferences, we apply DPO to align the model toward outputs with higher realism, stronger semantic consistency, and better controllability.
Extensive experiments show that VersaVogue consistently outperforms existing methods in both garment generation and virtual dressing, achieving superior image realism and fine-grained controllable precision. Our contributions are summarized as follows:
\begin{itemize}
    \item We propose VersaVogue, a unified fashion synthesis framework that seamlessly bridges garment generation and virtual dressing under a common multi-condition controllable formulation.
    
    \item We introduce a trait-routing attention (TA) module based on mixture-of-experts routing, which dynamically orchestrates heterogeneous visual conditions and enables precise disentanglement and adaptive attribute injection.
    
    \item We devise a multi-perspective preference optimization (MPO) strategy that automatically constructs high-quality preference data from content, semantic, and aesthetic evaluations, and further improves realism and controllability via direct preference optimization.
\end{itemize}

\begin{figure*}
    \centering
    \includegraphics[width=\linewidth]{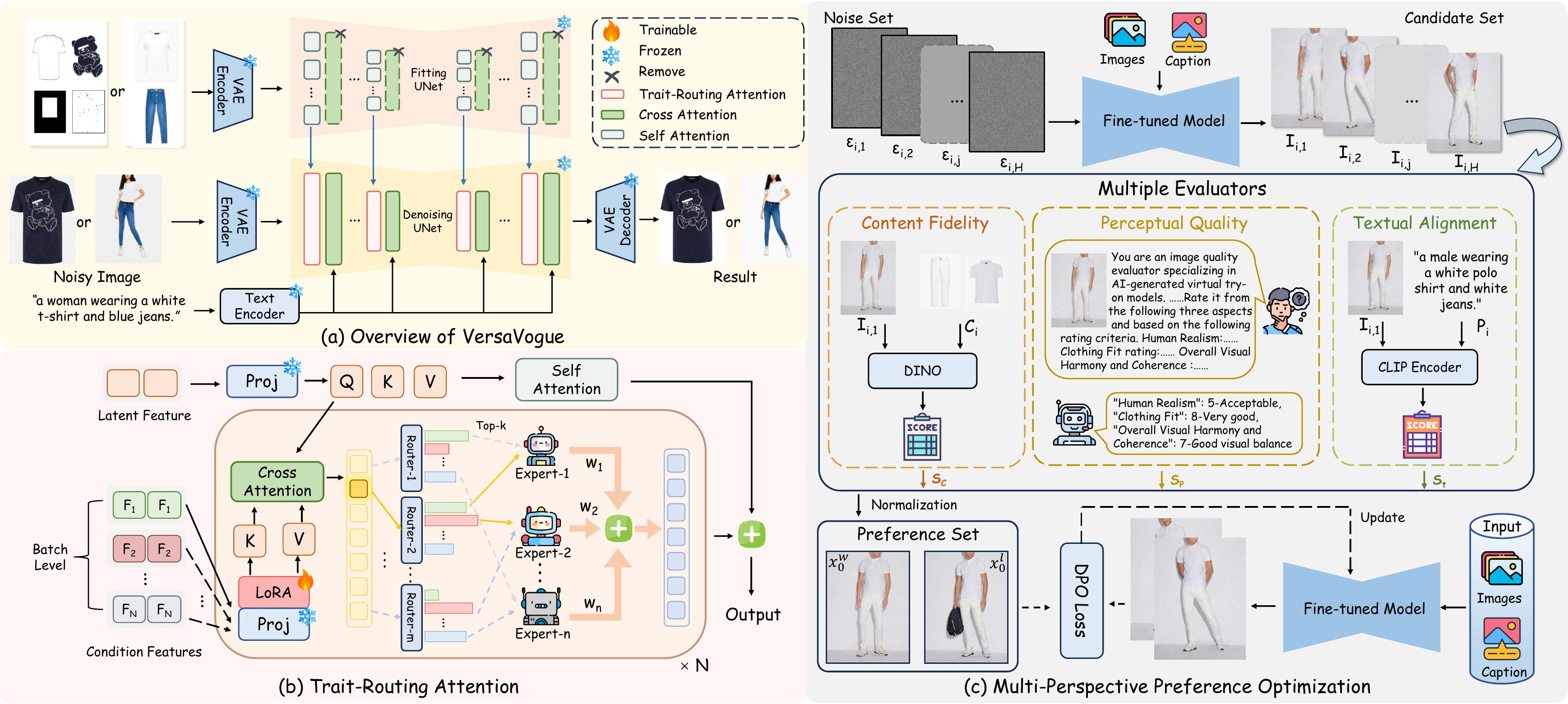}
    \caption{Architectural overview of VersaVogue. (a) First, distinct conditions are processed through isolated self-attention layers in the fitting UNet for independent feature extraction. (b) Then, we introduce a trait-routing attention (TA) module, utilizing a mixture-of-experts mechanism for dynamic feature fusion and injection into the denoising UNet. (c) Furthermore, we leverage multi-perspective preference optimization (MPO) strategy to automatically construct preference datasets and apply DPO loss to further optimize the model for generating results with higher fidelity and coherence.}
    \label{fig:framework}
\end{figure*}

\section{Related Work}

\noindent\textbf{Controllable Fashion Generation.} Recent advancements in diffusion models have spurred the development of adapters for controllable synthesis. General frameworks like ControlNet~\cite{zhang2023adding} and IP-Adapter~\cite{ye2023ip} introduce spatial or semantic guidance but often lack the fidelity required for fine-grained fashion tasks. To address domain-specific demands, methods such as ARMANI~\cite{zhang2022armani} and AnyLogo~\cite{zhang2024anylogo} align generation with structural sketches or logos, while IMAGGarment~\cite{shen2025imaggarment} employs multi-stage processing for attribute consistency. Similarly, virtual dressing approaches~\cite{chen2024magic,lin2025dreamfit} utilize fine-tuned UNets or attention injection to preserve garment details. Despite these progresses, existing methods typically treat multiple conditions via static stacking or simple concatenation. They lack adaptive mechanisms to dynamically disentangle heterogeneous inputs, leading to attribute entanglement and semantic interference in complex, multi-condition scenarios.

\noindent\textbf{Mixture of Experts.} Mixture of experts (MoE)~\cite{shazeer2017outrageously,fedus2022switch,fei2024scaling} scales model capacity efficiently by dynamically routing input tokens to specialized sub-networks ("experts") without increasing inference cost. In generative modeling, recent works have adapted MoE to manage diverse generation requirements. For instance, UIR-LoRA~\cite{zhang2024uir} treats LoRA layers as experts for varying image degradations, while RAPHAEL~\cite{xue2023raphael} employs spatial routing for regional image refinement. Drawing inspiration from these architectures, we integrate a token-level MoE strategy into our conditional injection mechanism. Unlike prior works focusing on task switching or regional editing, our approach leverages MoE to dynamically align specific visual attributes with the most compatible model layers, ensuring precise feature disentanglement during synthesis.

\noindent\textbf{Direct Preference Optimization.} Direct preference optimization (DPO)~\cite{rafailov2023direct} effectively aligns models with human intent by optimizing policy directly on preference data, bypassing explicit reward modeling. This paradigm has been successfully extended from LLMs to diffusion models by methods like DiffusionDPO~\cite{wallace2024diffusion} and D3PO~\cite{yang2024using}, which reformulate the denoising process for preference learning. In the fashion domain, FashionDPO~\cite{yu2025fashiondpo} utilizes user feedback to optimize outfit compatibility. However, most existing approaches rely on single-dimension metrics. In contrast, VersaVogue introduces a multi-perspective evaluation framework, encompassing content fidelity, textual alignment, and perceptual quality. By constructing a high-confidence preference dataset based on these comprehensive criteria, we employ DPO to rigorously enforce both semantic controllability and visual realism.

\section{Methodology}

\noindent\textbf{Problem Definition.} 
We formulate garment synthesis and virtual dressing as a unified multi-condition generative task, where the objective is to synthesize high-fidelity fashion images guided by heterogeneous visual priors and linguistic directives. Mathematically, this formulation is expressed as: 
\begin{equation}
    I_{\text{gen}} = \mathcal{G}(\mathcal{C}, P),
\end{equation}
where $\mathcal{G}$ represents our proposed generative model and $I_{\text{gen}}$ denotes the synthesized output. The conditioned input $\mathcal{C} = \{c_i\}_{i=1}^N$ aggregates a diverse ensemble of visual constraints: in the context of garment generation, $\mathcal{C}$ encapsulates the structural silhouette, chromatic hints, and a localized logo associated with a spatial mask; for virtual dressing, $\mathcal{C}$ comprises a set of discrete clothing exemplars. Here, $N$ represents the number of visual constraints, while $P$ acts as the linguistic instruction for high-level semantic control.

\noindent\textbf{Architecture Overview.} 
To achieve high-fidelity and controllable fashion synthesis, we propose two key modules: trait-routing attention (TA) and multi-perspective preference optimization (MPO), as shown in Figure~\ref{fig:framework}.
The trait-routing attention module leverages a mixture-of-experts framework to orchestrate autonomous layer-wise alignment. Specifically, it initiates dense interactions between visual conditions and latent embeddings via cross-attention, where the multi-expert mechanism dynamically distills the most compatible attribute features for each SDXL~\cite{podell2023sdxl} layer. This process establishes precise semantic guidance for the corresponding hierarchy.
Furthermore, we implement a dual-phase training strategy to optimize the model. Following an initial MSE-based phase that establishes a robust generative foundation, we introduce multi-perspective preference optimization to elevate perceptual fidelity. This involves performing stochastic sampling to populate a diverse candidate set, which is then rigorously audited by our multi-evaluator system. By discerning superior and inferior instances, we formulate a high-quality preference dataset to strictly align the model's policy with human aesthetics via the DPO loss.

\subsection{Trait-Routing Attention}
Drawing upon insights from prior literature~\cite{li2025anydressing}, it has been established that isolating self-attention layers while sharing parameters in other layers is an effective strategy to mitigate information interference when handling multiple visual inputs. Motivated by this premise, we employ a unified fitting UNet architecture equipped with dedicated attention branches to process $N$ distinct conditions simultaneously. Specifically, given $N$ visual condition images, we first project them into the latent space using a frozen VAE encoder. To facilitate efficient parallel processing, the resulting latent feature maps are concatenated along the batch dimension before being fed into the fitting UNet. As shown in Figure~\ref{fig:framework} (a), in the self-attention layers, we explicitly partition the concatenated features into distinct groups to perform independent attention calculations for each condition. This mechanism ensures that the semantic integrity of each condition is preserved. Formally, the operation is defined as:
\begin{equation}
    F_i^{new} = \text{Softmax}\left(\frac{\hat{Q}_i \hat{K}_i^T}{\sqrt{d}}\right) \hat{V}_i, \quad \forall i \in \{1, \dots, N\}.
\end{equation}
Formally, the projections are computed as $\hat{Q}_i = F_i (\hat{W}_q + \Delta \hat{W}_q^i)$, $\hat{K}_i = F_i (\hat{W}_k + \Delta \hat{W}_k^i)$, and $\hat{V}_i = F_i (\hat{W}_v + \Delta \hat{W}_v^i)$. In this formulation, $F_i$ denotes the input feature representation for the $i$-th condition. The weight matrices are decomposed into a frozen component $\hat{W}_{\{q,k,v\}}$, initialized from the pre-trained SDXL, and a learnable LoRA-based residual component $\Delta \hat{W}_{\{q,k,v\}}^i$, designed to facilitate attribute-specific adaptation. The resulting refined features $F_i^{new}$ are subsequently concatenated along the batch dimension and propagated into the denoising UNet. Additionally, given that the fitting UNet operates independently of textual conditioning, we eliminate the cross-attention layers to streamline the architecture.

Prior methods~\cite{frenkel2024implicit,wang2024instantstyle} have investigated the correlation between specific layers of SDXL~\cite{podell2023sdxl} and distinct visual attributes by employing selective injection strategies to mitigate interference. These approaches typically operate under a "disentangled reference" assumption where a single reference image dictates a singular attribute, such as utilizing one image exclusively for style transfer and another for chromatic guidance. However, such static assignment heuristics prove inadequate for our task where a single reference image imposes multiple entangled constraints like structure, texture, and chromaticity simultaneously. Consequently, precise manual injection into specific layers is infeasible. To address this, we propose the trait-routing attention module designed to facilitate the adaptive injection of these multi-faceted attributes. 
Formally, given the denoising latent features $Z$ and the set of condition features extracted by the fitting UNet denoted as $\{F_i\}_{i=1}^N$, we treat each condition as an independent processing stream. As depicted in Figure~\ref{fig:framework} (b), for each stream, we first execute a cross-attention operation between $Z$ and $F_i$ yielding an intermediate representation $U_i \in \mathbb{R}^{m \times d}$. This interaction is defined as:
\begin{equation}
    U_i =  \text{Softmax}\left(\frac{Q K_i^T}{\sqrt{d}}\right)V_i, \quad \forall i \in \{1, \dots, N\},
\end{equation}
where $Q = Z W_q$, $K_i = F_i(W_k + \Delta W_k^i)$, $V_i = F_i(W_v + \Delta W_v^i)$.
Here, $W_{\{q,k,v\}}$ denote the frozen projection matrices initialized from the pre-trained SDXL backbone, while $\Delta W_k^i$ and $\Delta W_v^i$ represent the trainable LoRA-based parameters.
To achieve layer-adaptive feature extraction, we introduce a token-wise mixture-of-experts mechanism. Specifically, for every token $x \in \mathbb{R}^d$ within the intermediate feature $U_i$, we employ a lightweight gating network to predict the routing weights. 
To prevent routing collapse and encourage expert exploration, we adopt the noisy top-$k$ gating strategy~\cite{shazeer2017outrageously}. The routing logits $H(x)$ are formulated as:
\begin{equation}
    H(x) = W_g x + \epsilon \cdot \text{Softplus}(W_{noise} x),
\end{equation}
where $W_g$ and $W_{noise}$ are trainable weight matrices, and $\epsilon \sim \mathcal{N}(0, 1)$ represents standard Gaussian noise. 
We then select the indices $\mathcal{T}$ corresponding to the top-$k$ values of $H(x)$, and the final routing weights $R$ are obtained by applying softmax to these selected logits:
\begin{equation}
    R_i = \left\{
    \begin{array}{c@{}l@{\quad}l} 
        \displaystyle \frac{e^{H(x)_i}}{\sum_{j \in \mathcal{T}} e^{H(x)_j}} & , & \text{if } i \in \mathcal{T} \\ [1em]
        0 & , & \text{otherwise}.
    \end{array}
    \right.
\end{equation}
Each expert $E_j$ is implemented as a Multilayer Perceptron (MLP) designed to extract features pertinent to specific attributes. The final output $x'$ is obtained via the weighted aggregation of the selected experts' outputs:
\begin{equation}
    x' = \sum_{j \in \mathcal{T}} R_j \cdot E_j(x).
\end{equation}
Subsequently, these individually processed tokens are concatenated to form the updated attribute features $F'_i$. Finally, these refined features are added to the result of the self-attention performed on $Z$, yielding the final output:
\begin{equation}
    Z_{out} = \text{Softmax}\left(\frac{Q K^T}{\sqrt{d}}\right) V + \sum_{i}^N F'_i,
\end{equation}
where $K=Z W_k$, $V = Z W_v$. This trait-routing attention module ensures dynamic alignment between the injected attribute features and the model layers, realizing an adaptive injection of attribute features that is compatible with layer-specific representations.

\noindent \textbf{Training.} 
Prior to the subsequent optimization, we conduct preliminary training to obtain an initial virtual dressing model, denoted as $\epsilon_{ref}$. During this training phase, we optimize the model using the mean squared error (MSE) loss, consistent with the standard objective of latent diffusion models~\cite{rombach2022high}: 
\begin{equation}
    \mathcal{L} = \mathbb{E}_{z_0,\epsilon,\mathcal{C},P,t}\| \epsilon - \epsilon_{\mathcal{G}}(z_t,\mathcal{C},P,t) \|^2.
\label{eq:mse_loss}
\end{equation}
\begin{figure*}[t]
    \centering
    \includegraphics[width=0.98\linewidth]{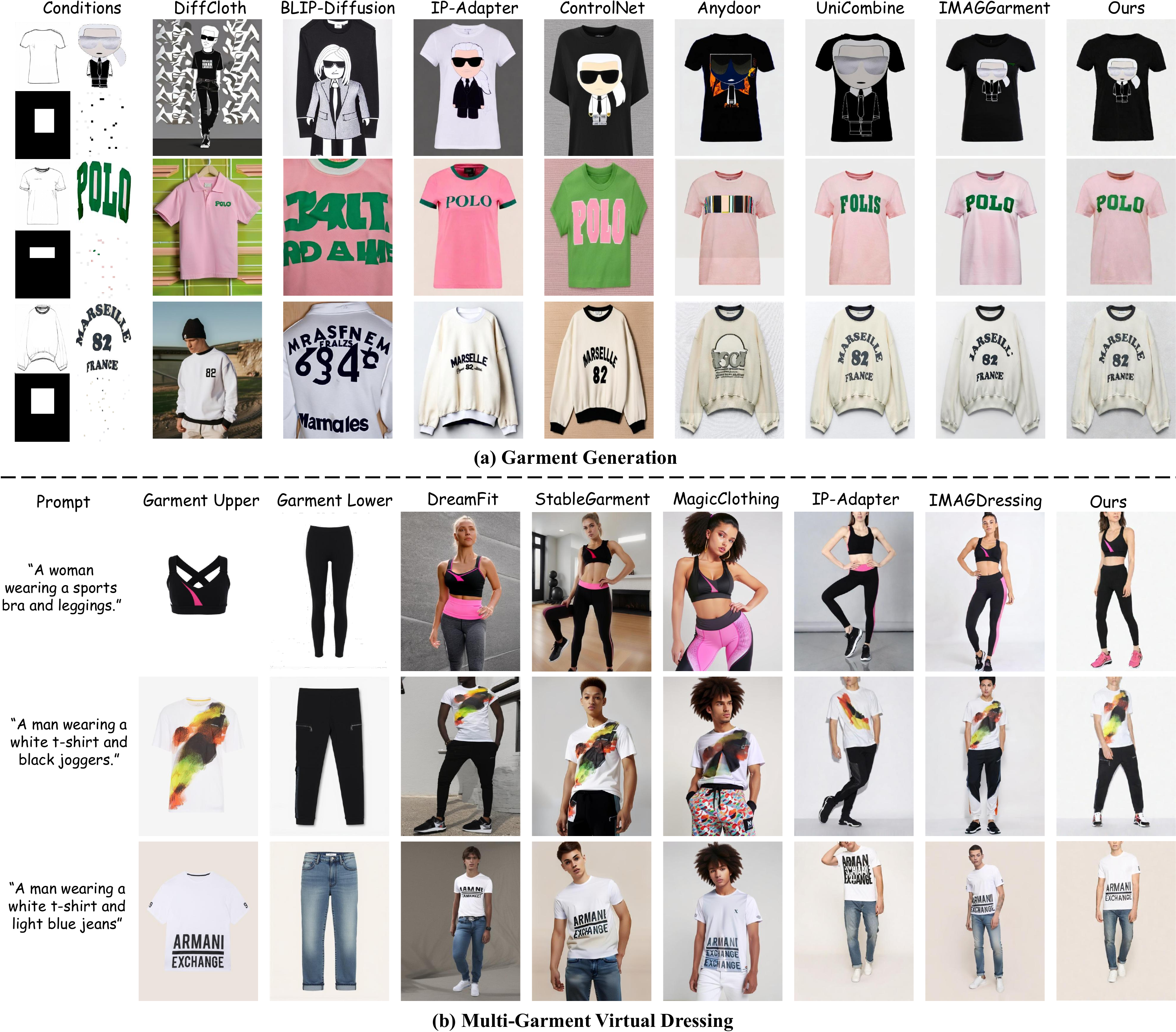}
    \caption{Qualitative comparison with baseline methods on (a) garment generation and (b) multi-garment virtual dressing. Our method demonstrates superior realism and conditional controllability. }
    \label{fig:gg_and_multi-garment}
    \vspace{-0.4cm}
\end{figure*}

\subsection{Multi-Perspective Preference Optimization}

Following the initial training stage, we derive a preliminary virtual dressing model $\epsilon_{ref}$. While functional, it falls short of delivering optimal photorealism and precise alignment with strict visual constraints. To bridge the gap towards high-fidelity synthesis and precise condition adherence, we introduce a multi-perspective preference optimization strategy. Specifically, we first curate a sampling dataset $\mathcal{D} = \{(\mathcal{C}_i, P_i)\}_{i=1}^M$ by drawing $M$ instances from the training distribution. For each sample, we introduce stochastic diversity by sampling $H$ random noise images, forming the noise set $ \{\{\epsilon_{i,j}\} _{j=1}^H\}_{i=1}^M$, prompting the frozen reference model $\epsilon_{ref}$ to synthesize a diverse pool of candidate images $\{\{I_{i,j}\}_{j=1}^H\}_{i=1}^M$. To transcend the limitations of single-metric assessment, these candidates are scrutinized through a robust multi-evaluator framework that integrates feedback on content fidelity, perceptual quality, and textual alignment. Guided by this holistic critique, we distill high-confidence preference pairs to steer the subsequent optimization. We elaborate on the design of these evaluators below:

 \noindent\textbf{(1) Content Fidelity.} 
The alignment between synthesized imagery and user-specified visual conditions constitutes a critical measure of a model's controllability. To guarantee the preservation of fine-grained attributes, we introduce the content fidelity evaluator, which leverages the DINO~\cite{zhang2022dino} model for similarity computation. This allows us to effectively examine local semantic correspondences and verify the structural consistency of the generated content. Specifically, for each candidate image $I_{i,j}$ and its corresponding set of visual conditions $\mathcal{C}_i$, we project both inputs into a shared latent space via the pre-trained DINO encoder $\phi_{\text{DINO}}$. We then derive the content fidelity score by computing the arithmetic mean of the pairwise cosine similarities between the candidate and each condition embedding, as formulated below:
\begin{equation}
    s_{i,j}^c = \frac{1}{|\mathcal{C}_i|} \sum_{c \in \mathcal{C}_i} \text{CosSim}\left( \phi_{\text{DINO}}(I_{i,j}), \phi_{\text{DINO}}(c) \right).
\end{equation}

\noindent \textbf{(2) Perceptual Quality.}
Traditional quantitative evaluation metrics, exemplified by Fréchet Inception Distance (FID)~\cite{heusel2017gans} and Structural Similarity Index (SSIM)~\cite{wang2004image}, typically rely on calculating distributional discrepancies within low-level feature spaces. Consequently, these methods often fail to capture high-level semantic nuances or align with human aesthetic perception. To bridge this gap, we employ the multi-modal large language model CogVLM~\cite{hong2024cogvlm2}, denoted as $\mathcal{M}$, to act as a scalable surrogate for human judgment. Specifically, we feed each candidate image into $\mathcal{M}$ along with a tailored instruction prompt that directs the model to critically assess the visual content from three pivotal dimensions: visual photorealism, structural rationality, and aesthetic harmony. To facilitate quantitative analysis, we instruct the model to output a discrete numerical score ranging from 1 to 10, where 1 indicates severe degradation and 10 denotes optimal quality. By parsing the structured textual feedback and aggregating the distinct sub-scores, we derive the final perceptual quality score $s_{i,j}^p$ for each candidate.

\noindent \textbf{(3) Textual Alignment.}
Textual prompts dictate the overarching semantic context of the generation, serving as the primary directive for the image's content. To guarantee strict semantic consistency with these input instructions, we employ the pre-trained CLIP~\cite{radford2021learning} model as our evaluator. By projecting both the synthesized image and the text prompt into a shared embedding space, we can quantify their semantic correlation. Specifically, given a candidate image $I_{i,j}$ and the corresponding textual prompt $P_i$, we map both inputs into a unified latent space using the CLIP image encoder $\phi_{\text{img}}$ and text encoder $\phi_{\text{txt}}$. The textual alignment score $s_{i,j}^t$ is then computed as the cosine similarity between their normalized embeddings:
\begin{equation}s_{i,j}^t = \text{CosSim}\big( \phi_{\text{img}}(I_{i,j}), \phi_{\text{txt}}(P_i) \big).\end{equation}

\noindent\textbf{Composite Control Alignment.} 
Diffusion Direct Preference Optimization~\cite{wallace2024diffusion} (DPO) serves as a potent alignment strategy, steering the generative distribution towards high-fidelity regions while repelling it from sub-optimal modes, thereby ensuring the generated imagery aligns with human perception. To enhance the realism and controllability of our model, we employ DPO to refine the weights based on a preference dataset constructed via our multi-evaluator framework. 
Specifically, given a conditioning input $(\mathcal{C}_i, P_i)$, we generate a set of candidate images $\{I_{i,j}\}_{j=1}^H$. We evaluate these candidates across the three dimensions, yielding the corresponding raw score vectors $\mathcal{S}_i^c = \{ s_{i,j}^c\}_{j=1}^H, \mathcal{S}_i^p= \{ s_{i,j}^p\}_{j=1}^H, \mathcal{S}_i^t= \{ s_{i,j}^t\}_{j=1}^H$. To mitigate scale discrepancies among different metrics, we normalize these raw scores before aggregation. The total score set $\mathcal{S}_{i}^{total}$ is obtained by the element-wise summation of the normalized vectors:
\begin{equation}
\mathcal{S}_{i}^{total} = \text{Norm}(\mathcal{S}_{i}^c) + \text{Norm}(\mathcal{S}_{i}^p) + \text{Norm}(\mathcal{S}_{i}^t) ,
\end{equation}
where $\text{Norm}(\cdot)$ denotes the Z-score normalization across the candidate dimension. Based on this total score, the image with the highest score is selected as the winning sample $x_0^w$, while the lowest-scoring image serves as the losing sample $x_0^l$. Subsequently, we optimize the model using the DPO objective, which implicitly maximizes the margin between the likelihoods of the preferred and dispreferred distributions. The objective $\mathcal{L}_{DPO}$ is defined as follows:
\begin{equation}
\begin{aligned}
    \mathcal{L}_{DPO} &= -\mathbb{E}_{x^w_0, x^l_0, \mathcal{C}^\prime\sim \mathcal{D}, t, \epsilon}   \log \sigma \Big( -\beta T \omega(\lambda_t) \Big[ \\
    &\quad \left( \|\epsilon - \epsilon_\theta(x_{t}^w, \mathcal{C}^\prime, t) \|_2^2 - \|\epsilon - \epsilon_{ref}(x_{t}^w, \mathcal{C}^\prime, t) \|_2^2 \right) - \\
    &\quad \left( \|\epsilon - \epsilon_\theta(x_{t}^l, \mathcal{C}^\prime, t) \|_2^2 - \|\epsilon - \epsilon_{ref}(x_{t}^l, \mathcal{C}^\prime, t) \|_2^2 \right) \Big] \Big). 
\end{aligned}
\label{eq:dpo}
\end{equation}
Here, $\mathcal{C}^\prime = \{\mathcal{C}, P\}$ encompasses all input conditions, $t$ represents the timestep, and $\epsilon$ denotes the sampled Gaussian noise. $\beta$ serves as a hyperparameter, $\lambda_t$ signifies the signal-to-noise ratio, and $\omega(\lambda_t)$ acts as a weighting function that is fixed in practice. The trainable model $\epsilon_\theta$ is initialized from the reference model $\epsilon_{\text{ref}}$, which remains frozen during training. This optimization strategy effectively aligns the model with the preferred sample $x_0^w$, thereby steering the generative process toward superior image quality. 
\begin{table}[t]
    \centering
    \caption{Quantitative comparison with state-of-the-art methods for garment generation on the GarmentBench dataset.}
    \label{tab:gg_comparsion}

    \setlength{\tabcolsep}{5pt}
    \renewcommand{\arraystretch}{1.1}

    \resizebox{\columnwidth}{!}{
    \begin{tabular}{l cccc}
        \toprule
        \textbf{Method} & LLA↑ & CSS↓ & FID↓ & LPIPS↓ \\
        \midrule
        
        DiffCloth~\cite{zhang2023diffcloth} & 0.153 & 108.199 & 137.973 & 0.605 \\
        BLIP-Diffusion~\cite{li2023blip} & 0.127 & 104.439 & 106.846 & 0.676 \\
        IP-Adapter~\cite{ye2023ip} & 0.268 & 98.755 & 43.208 & 0.417 \\
        ControlNet~\cite{zhang2023adding} & 0.267 & 103.477 & 55.201 & 0.521 \\
        
        AnyDoor~\cite{chen2024anydoor} & 0.649 & 68.242 & 40.899 & 0.173 \\
        
        UniCombine~\cite{wang2025unicombine} 
        & 0.635 
        & \cellcolor{yellow!15} 47.391 
        & 38.210 
        & 0.137 \\
        
        IMAGGarment~\cite{shen2025imaggarment} 
        & \cellcolor{yellow!15}0.765 
        & 57.292 
        & \cellcolor{orange!20}16.514 
        & \cellcolor{yellow!15}0.116 \\
        
        \midrule

        Ours 
        & \cellcolor{orange!20}0.864 
        & \cellcolor{orange!20}41.719 
        & \cellcolor{yellow!15}17.060 
        & \cellcolor{orange!20}0.090 \\
        
        \bottomrule
    \end{tabular}
    }
\end{table}

\subsection{Training and Inference}
In the training phase, we exclusively optimize the added LoRA layers, as well as the routing networks and expert modules within the trait-routing attention. The training pipeline consists of two sequential stages. In the initial stage, we optimize the model using MSE loss formulated in Eq.~\ref{eq:mse_loss}.
Subsequently, we further refine the model to align with human aesthetic standards and semantic fidelity. We leverage our proposed multi-perspective preference optimization, optimizing the model via the DPO loss defined in Eq.~\ref{eq:dpo}.
During the inference stage, we employ classifier-free guidance (CFG)~\cite{ho2022classifier} to reinforce the controllability of the generation process. The adjusted noise prediction is defined as:
\begin{equation}
\hat{\epsilon}(z_t,\mathcal{C},P,t) = \omega \cdot \epsilon_{\mathcal{G}}(z_t,\mathcal{C},P,t) + (1-\omega) \cdot \epsilon_{\mathcal{G}}(z_t,t),
\end{equation}
where $\omega$ represents the guidance scale, governing the intensity of the conditional signal relative to the unconditional baseline.

\begin{figure}[t]
    \centering
    \includegraphics[width=\linewidth]{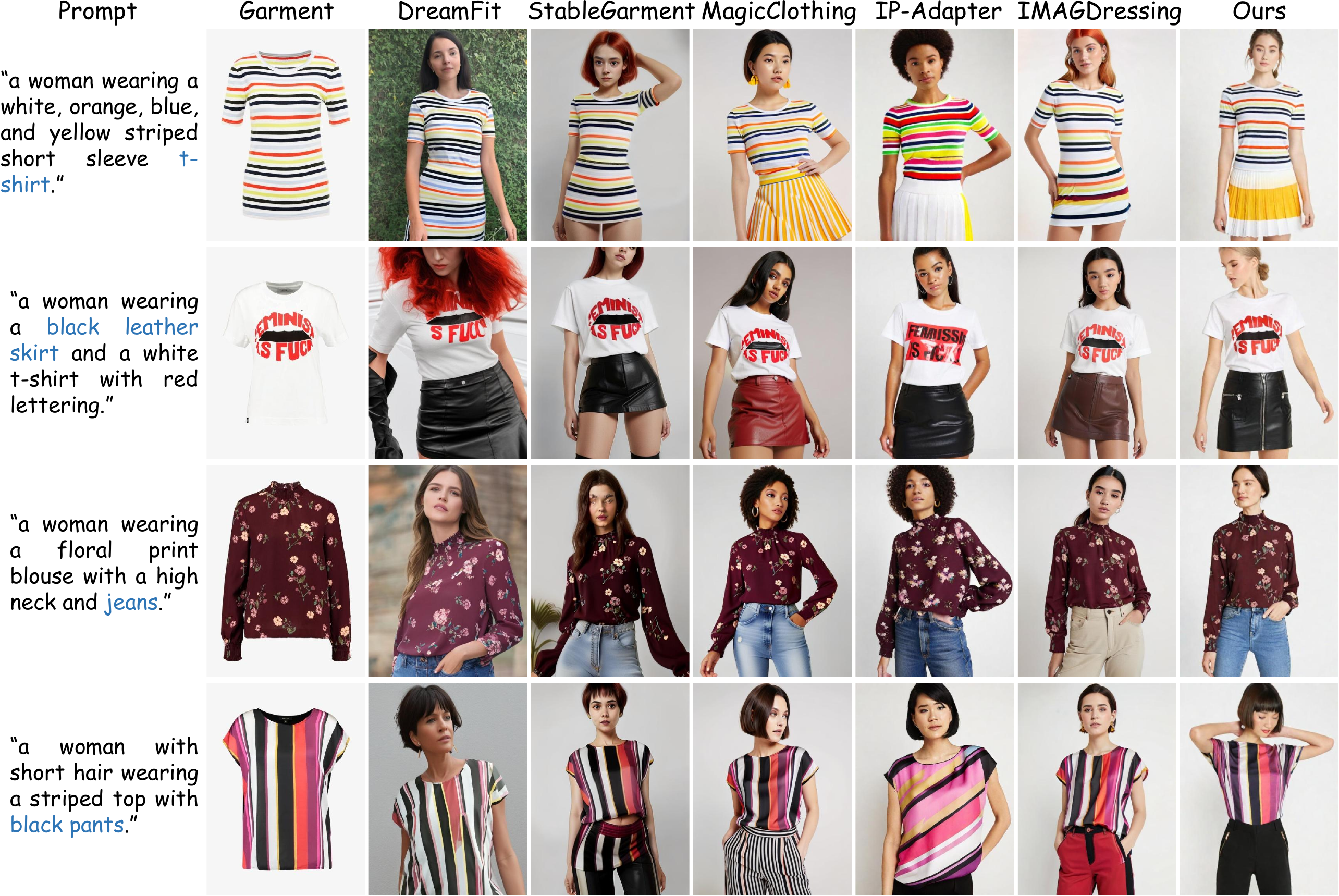}
    \caption{Qualitative comparison with SOTA methods on single-garment virtual dressing. Our method generates images with superior text alignment and detail preservation.}
    \label{fig:single-garment}
    \vspace{-0.4cm}
\end{figure}

\section{Experiments}
\subsection{Implementation Details}
\noindent \textbf{Dataset.}
We train our proposed framework using a comprehensive collection of public benchmarks tailored to distinct generative tasks. For garment generation, we adopt GarmentBench~\cite{shen2025imaggarment}, which comprises 8,200 training and 500 testing samples annotated with multi-aspect attributes. For single-garment virtual dressing, we employ VITON-HD~\cite{choi2021viton} and DressCode~\cite{morelli2022dresscode}, containing 11,647 and 48,392 high-resolution training pairs, respectively.
To address complex multi-garment scenarios, we curate a specialized subset from DressCode-MR~\cite{chong2025fastfitacceleratingmultireferencevirtual} by filtering exclusively for instances with paired upper and lower garments, resulting in 7,848 training entries. Since these original dressing benchmarks lack textual descriptions, we leverage Qwen-VL-chat~\cite{Qwen-VL} to automatically caption the images, thereby enabling our text-guided synthesis.
For evaluation, we sample 300 pairs from the official test splits, encompassing diverse colors, types, and styles to comprehensively assess performance.

\begin{table*}[t]
    \centering
    \caption{Quantitative comparison with SOTA methods for virtual dressing, evaluated on VITON-HD and DressCode for single-garment, and on DressCode-MR for multi-garment.}
    \label{tab:comparison}
    \small

    \setlength{\tabcolsep}{4.0pt}
    \renewcommand{\arraystretch}{1.08}
    \setlength{\fboxsep}{3.8pt}

    \resizebox{0.98\textwidth}{!}{
    \begin{tabular}{l | cccc | cccc | cccc}
        \toprule
        \multirow{2}{*}{Method} 
        & \multicolumn{8}{c|}{\textbf{Single-Garment}} 
        & \multicolumn{4}{c}{\textbf{Multi-Garment}} \\ 
        
       \cline{2-9} \cline{10-13}
        
        & \multicolumn{4}{c|}{VITON-HD~\cite{choi2021viton}} 
        & \multicolumn{4}{c|}{DressCode~\cite{morelli2022dresscode}} 
        & \multicolumn{4}{c}{DressCode-MR~\cite{chong2025fastfitacceleratingmultireferencevirtual}} \\
        
        \cline{2-13} 
        
        & FID↓ & CLIP-I↑ & SSIM↑ & LPIPS↓
        & FID↓ & CLIP-I↑ & SSIM↑ & LPIPS↓
        & FID↓ & CLIP-I↑ & SSIM↑ & LPIPS↓ \\
        \midrule
        
        DreamFit~\cite{lin2025dreamfit}     
        & 79.527 & 0.666 & 0.439 & 0.611 
        & 87.256 & 0.743 & 0.479 & 0.591 
        & 114.785 & 0.663 & 0.470 & 0.626 \\
        
        StableGarment~\cite{wang2024stablegarment}       
        & 57.876 & 0.762 & 0.609 & 0.441   
        & 79.495 & 0.795 & 0.629 & 0.442 
        & 104.428 & 0.753 & 0.610 & 0.485 \\
        
        MagicClothing~\cite{chen2024magic}  
        & 53.113 & 0.749 & 0.643 & 0.405  
        & 105.320 & 0.770 & 0.628 & 0.460 
        & 134.816 & 0.746 & 0.661 & 0.444 \\
        
        IP-Adapter~\cite{ye2023ip}    
        & 50.033 & 0.770 & 0.632 & 0.422  
        & 62.615 & 0.790 & 0.640 & 0.434 
        & \cellcolor{yellow!25}{50.600} & 0.789 & 0.762 & 0.305 \\
        
        IMAGDressing~\cite{shen2025imagdressing} 
        & \cellcolor{yellow!25}43.210 & \cellcolor{yellow!25}0.796 & \cellcolor{yellow!25}{0.676} & \cellcolor{orange!25}{0.363}  
        & \cellcolor{yellow!25}{59.827} & \cellcolor{orange!25}{0.805} & \cellcolor{yellow!25}{0.722} & \cellcolor{yellow!25}{0.349} 
        & 56.380 & \cellcolor{yellow!25}{0.792} & \cellcolor{yellow!25}{0.775} & \cellcolor{yellow!25}{0.296} \\
        \midrule
        
        Ours  
        & \cellcolor{orange!25}{39.772} & \cellcolor{orange!25}{0.809} & \cellcolor{orange!25}{0.685} & \cellcolor{yellow!25}{0.364}  
        & \cellcolor{orange!25}{53.648} & \cellcolor{yellow!25}{0.800} & \cellcolor{orange!25}{0.724} & \cellcolor{orange!25}{0.346} 
        & \cellcolor{orange!25}{44.740} & \cellcolor{orange!25}{0.809} & \cellcolor{orange!25}{0.811} & \cellcolor{orange!25}{0.225} \\
        \bottomrule
    \end{tabular}
    }
\end{table*}

\noindent \textbf{Hyper-Parameters.}
We initialize VersaVogue using SDXL v1.0 base model~\cite{podell2023sdxl} with a LoRA rank of 128, optimizing only the LoRA layers, along with the routers and experts within the TA module. We optimize the model using AdamW with a batch size of 1 on two NVIDIA A800 GPUs. The image resolutions are set to $768 \times 576$ for virtual dressing and $640 \times 512$ for garment generation. We set $n=4$ experts and a top-$k$ value of 2. The initial 400K-step training uses a learning rate of $1 \times 10^{-5}$. We then conduct a 40K-step multi-perspective preference optimization with $H=10$ and $M=1000$, adjusting the DPO learning rate to $8.192 \times 10^{-9}$ and $\beta=5000$. To facilitate CFG~\cite{ho2022classifier}, a $5\%$ dropout rate is applied to each condition during training. For inference, we employ the DDIM sampler~\cite{song2022denoisingdiffusionimplicitmodels} with $50$ steps and set the guidance scale $\omega$ to $7.5$.

\begin{figure}
    \centering
     \includegraphics[width=0.98\linewidth]{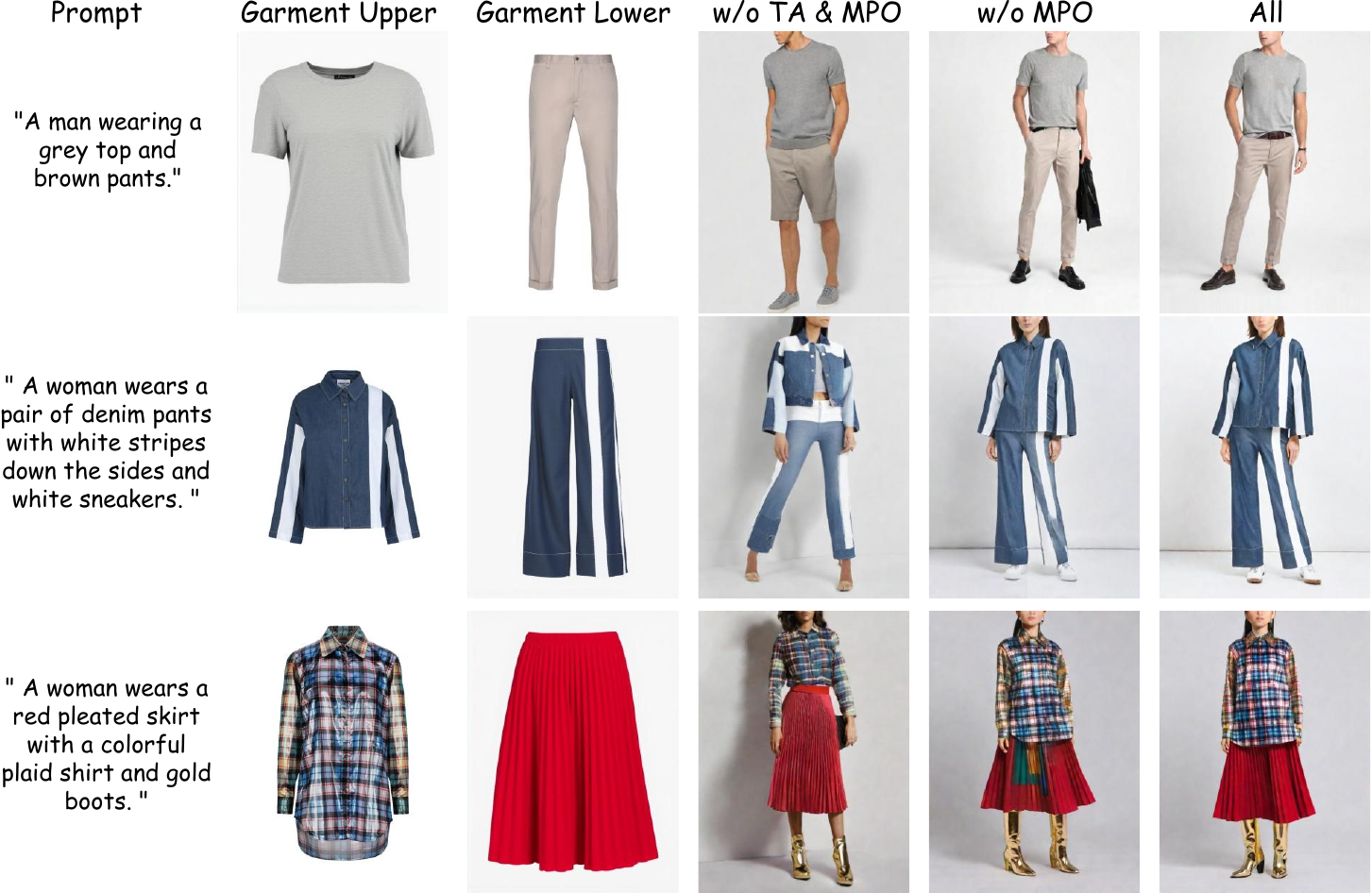}
    \caption{Qualitative results of ablation study.}
    \label{fig:ablation}
    \vspace{-0.5cm}
\end{figure}

\subsection{Qualitative Comparisons}

\noindent \textbf{Garment Generation.} We conduct extensive comparisons with seven relevant state-of-the-art approaches, including DiffCloth~\cite{zhang2023diffcloth}, BLIP-Diffusion~\cite{li2023blip}, IP-Adapter~\cite{ye2023ip}, ControlNet~\cite{zhang2023adding}, AnyDoor~\cite{chen2024anydoor}, UniCombine~\cite{wang2025unicombine}, and IMAGGarment~\cite{shen2025imaggarment}. Visual results in Figure~\ref{fig:gg_and_multi-garment} demonstrate that even with the logo treated as a specific subject for inpainting, AnyDoor and UniCombine exhibit limitations in maintaining local details for complex logos. While IMAGGarment employs a two-stage strategy to handle attribute diversity, it falls short in preserving fine-grained details and maintaining global visual harmony. Instead, by incorporating a dynamic routing mechanism tailored for distinct features, our method delivers superior performance in both detail reconstruction and overall visual fidelity.

\noindent \textbf{Virtual Dressing.}
We conduct a detailed comparison of our method against leading baselines in image generation, including DreamFit~\cite{lin2025dreamfit}, StableGarment~\cite{wang2024stablegarment}, MagicClothing~\cite{chen2024magic}, IP-Adapter~\cite{ye2023ip}, and IMAGDressing~\cite{shen2025imagdressing}. 
Due to the scarcity of publicly available methods dedicated to multi-garment virtual dressing, we also employ these baselines for our multi-garment comparison, obtaining comparable results by spatially concatenating features across multiple garments. 
We use the official model parameters from their implementations. Figure ~\ref{fig:single-garment} presents the qualitative comparison results of single-garment virtual dressing between our method and baselines. We observe that while MagicClothing and IMAGDressing produce visually plausible images, they exhibit noticeable deficiencies in preserving conditional consistency and semantic alignment with text prompts. 
Moreover, as depicted in Figure~\ref{fig:gg_and_multi-garment}, due to the lack of dedicated processing schemes for multi-garment inputs, these approaches fail to maintain the structural integrity and chromatic fidelity of the reference garments, particularly for complex color layouts and sophisticated logos, leading to suboptimal synthesis with significant detail distortion. Conversely, our method exhibits superior fidelity in restoring intricate details, yielding high-quality dressing results with substantial coherence and perceptual realism. 

\begin{table}[t]
    \centering
    \caption{Quantitative ablation on the DressCode-MR dataset.}
    \label{tab:ablation}

    \setlength{\tabcolsep}{5pt}
    \renewcommand{\arraystretch}{1.1}

    \setlength{\aboverulesep}{0pt}
    \setlength{\belowrulesep}{0pt}

    \resizebox{\columnwidth}{!}{

        \begin{tabular}{cc|cccc}
            \toprule
            \textbf{TA} & \textbf{MPO} & FID $\downarrow$ & CLIP-I $\uparrow$ & SSIM $\uparrow$ & LPIPS $\downarrow$ \\
            \midrule
            
            \xmark & \xmark & 51.220 & 0.777 & 0.754 & 0.310 \\

            \cmark & \xmark 
            & \cellcolor{yellow!15}45.319 
            & \cellcolor{yellow!15}0.805  
            & \cellcolor{yellow!15}0.802 
            & \cellcolor{yellow!15}0.244 \\

            \cmark & \cmark 
            & \cellcolor{orange!20}44.740 
            & \cellcolor{orange!20}0.809  
            & \cellcolor{orange!20}0.811 
            & \cellcolor{orange!20}0.225 \\
            
            \bottomrule
        \end{tabular}
    }
\end{table}
\begin{figure}
    \centering
    \includegraphics[width=0.98\linewidth]{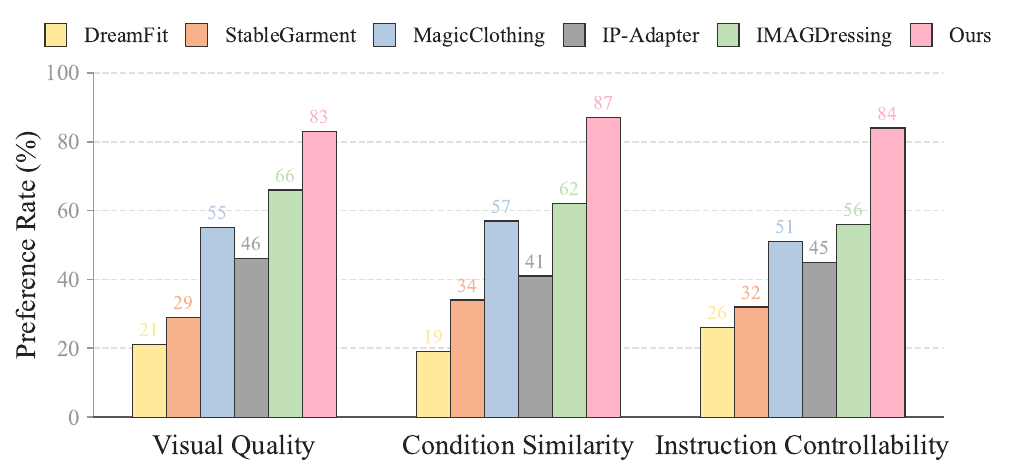}
    \caption{User study results. Higher scores indicate better visual realism and user preference.}
    \label{fig:user_study}
    \vspace{-0.5cm}
\end{figure}

\subsection{Quantitative Comparisons}
Table~\ref{tab:gg_comparsion} presents a quantitative comparison between our method and the baselines on the garment generation task. Adapter-based methods (\emph{e.g.}, IP-Adapter, ControlNet) process conditions in isolation due to the lack of mechanisms to model internal correlations, resulting in feature conflicts and suboptimal CSS and FID scores. Furthermore, as shown in Table~\ref{tab:comparison}, we conduct quantitative comparisons on the virtual dressing task across multiple datasets. Regarding single-garment evaluation, extensive experiments on the VITON-HD and DressCode datasets validate the competitiveness of our method. Moreover, owing to the limitations of naive fusion strategies, IMAGDressing and Dreamfit exhibit inferior CLIP-I and SSIM scores in multi-garment virtual dressing tasks. In contrast, our method outperforms baselines on all metrics on the multi-garment virtual dressing task, which further validates the effectiveness of our VersaVogue in achieving precise multi-condition control.

\subsection{Ablation Study}
To demonstrate the effectiveness of our proposed modules, we conduct ablation studies on the DressCode-MR dataset. We compare two variants: the first excludes both components, where conditional features independently interact with latents and are fused via simple summation; the second omits only the MPO strategy, directly employing the preliminary reference model $\epsilon_{ref}$ from the initial training stage. As illustrated in the third column of Figure~\ref{fig:ablation}, the absence of the TA module renders the model prone to severe entanglement between different attributes. This leads to information clutter, ultimately yielding inconsistent and low-quality synthetic images. The sharp decline in CLIP-I and SSIM values in Table~\ref{tab:ablation} underscores the efficacy of the TA module in decoupling distinct attributes and facilitating precise dynamic feature injection. Furthermore, Figure~\ref{fig:ablation} demonstrates that without MPO, the model tends to produce results with perceptual incoherence and degraded quality, failing to align with human aesthetic preferences. This validates that the MPO strategy is pivotal for enhancing the overall plausibility and realism of the generated images.

\subsection{More Analysis and Results}

\noindent \textbf{User Study.}
We conducted a user study where 20 participants selected their top-3 results across 20 random test cases based on three criteria: visual quality, condition similarity, and instruction controllability. As shown in Figure~\ref{fig:user_study}, the results consistently reveal that VersaVogue achieves the highest preference scores across all three metrics. In particular, participants favored VersaVogue for producing more realistic visuals while preserving fine-grained condition fidelity and maintaining stable control under heterogeneous inputs, which addresses the attribute entanglement and semantic interference commonly observed in prior methods. This human-aligned outcome supports our design goal of improving both perceptual realism through expert orchestration and preference alignment.

\noindent\textbf{Analysis of Token Flow.}
\begin{figure}
    \centering
    \includegraphics[width=0.98\linewidth]{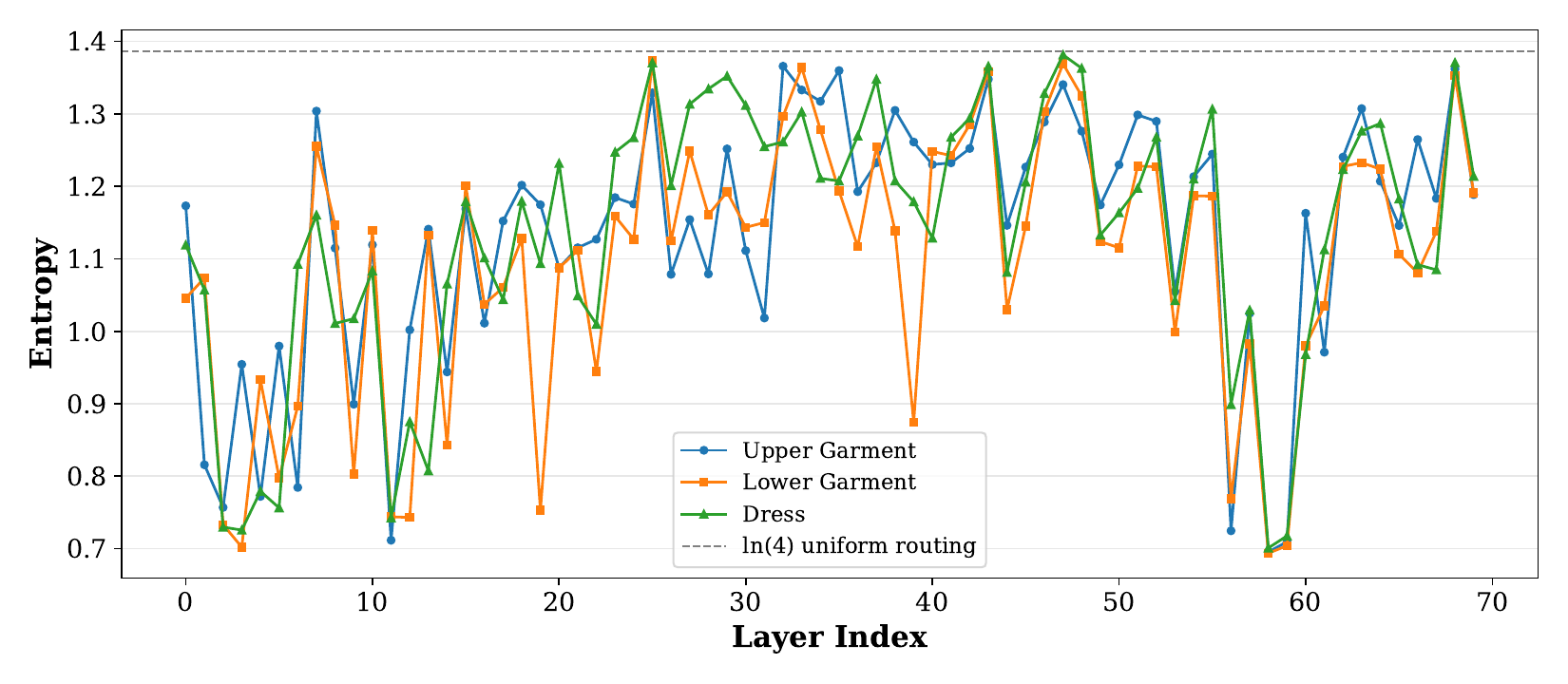}
    \caption{Dynamics of routing entropy across SDXL layers for distinct apparel categories.}
    \label{fig:entropy}
    \vspace{-0.5cm}
\end{figure}
To deeply investigate the internal decision-making mechanism of our proposed TA module during fashion image synthesis, we analyze the routing entropy dynamics across different layers of the SDXL. As illustrated in Figure~\ref{fig:entropy}, the routing entropy across all layers consistently remains below the theoretical upper bound of a uniform distribution ($\ln(4)$). The values exhibit varying degrees of dispersion at different network stages, indicating that the token allocation process is neither random nor evenly distributed among the available experts. Furthermore, the overall trajectories of these entropy changes show significant overlap across different garment categories. This structural consistency suggests that token allocation is fundamentally driven by the inherent characteristics of the garment attributes, enabling the model to dynamically adapt its routing strategy to specific input traits.

\noindent \textbf{Real-World Applications.}
\begin{figure}
    \centering
    \includegraphics[width=0.98\linewidth]{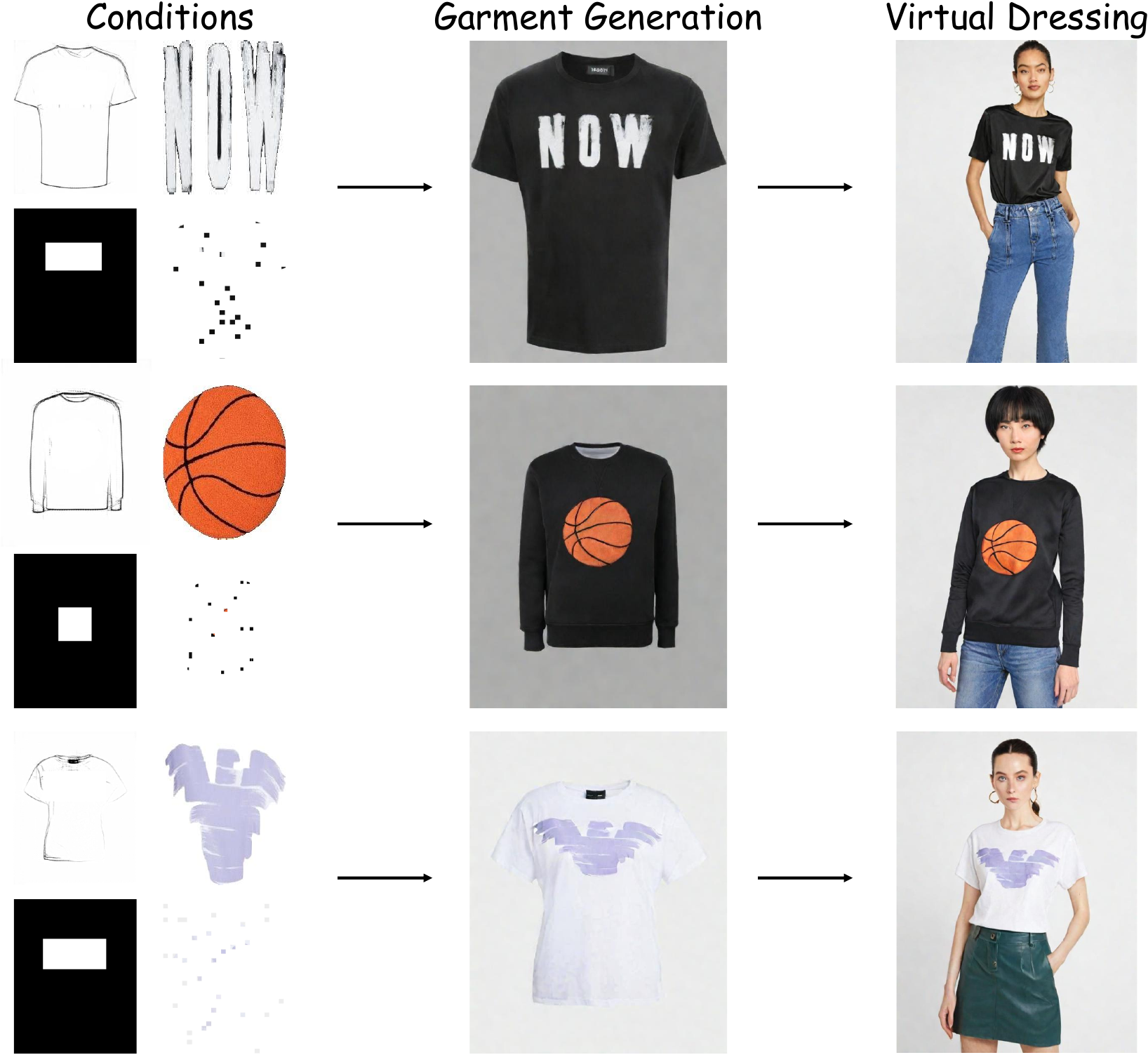}
    \caption{Real-world applications of VersaVogue, bridging garment generation and virtual dressing in one framework.}
    \label{fig:unite_result}
    \vspace{-0.6cm}
\end{figure}
Figure~\ref{fig:unite_result} demonstrates the application of our unified framework in real-world scenarios, successfully bridging the gap between garment generation and virtual dressing. It empowers users to freely customize their desired garments and instantly visualize the on-body effect without requiring any auxiliary reference inputs (\emph{e.g.}, human face or pose images). Furthermore, our model excels in maintaining strict semantic consistency with the input conditions and meticulously reconstructing fine-grained details to yield high-fidelity, photorealistic results. This seamless capability delivers a richer and more personalized experience for both online shoppers and e-commerce merchants.

\section{Conclusions}
In this paper, we propose VersaVogue, a unified framework tackling multi-condition garment generation and virtual dressing. We introduce a trait-routing attention (TA) module based on a token-level MoE mechanism, which adaptively routes distinct attribute features across SDXL layers for precise disentanglement. Furthermore, our multi-perspective preference optimization (MPO) strategy automates preference dataset construction by evaluating candidates on content fidelity, perceptual quality, and text alignment. Leveraging this dataset via direct preference optimization (DPO), VersaVogue significantly enhances perceptual realism and conditional controllability. Extensive experiments demonstrate that our method achieves state-of-the-art performance, outperforming existing approaches in visual fidelity and detail preservation.

\bibliographystyle{ACM-Reference-Format}
\bibliography{ref}

\clearpage
\appendix
\begin{center}
\section*{Supplementary Material}
\end{center}
The appendices provide additional details that support and extend
the main paper. 
Appendix~\ref{sec:more_results} provides further ablation studies and additional qualitative comparisons.
Appendix~\ref{sec:discussion} elucidates the underlying motivation for our model and provides an in-depth analysis of existing methods.
Appendix~\ref{sec:dpo_loss} provides the complete mathematical derivation of the DPO loss used in our MPO strategy.
Appendix~\ref{sec:more_details} delineates the perceptual quality prompts integrated into our MPO framework, accompanied by formal definitions of the evaluation metrics.
Finally, Appendix~\ref{sec:limitation} discusses the current limitations of VersaVogue and outlines directions for future extensions.

\section{More Results}
\label{sec:more_results}
\begin{figure}
    \centering
    \includegraphics[width=0.98\linewidth]{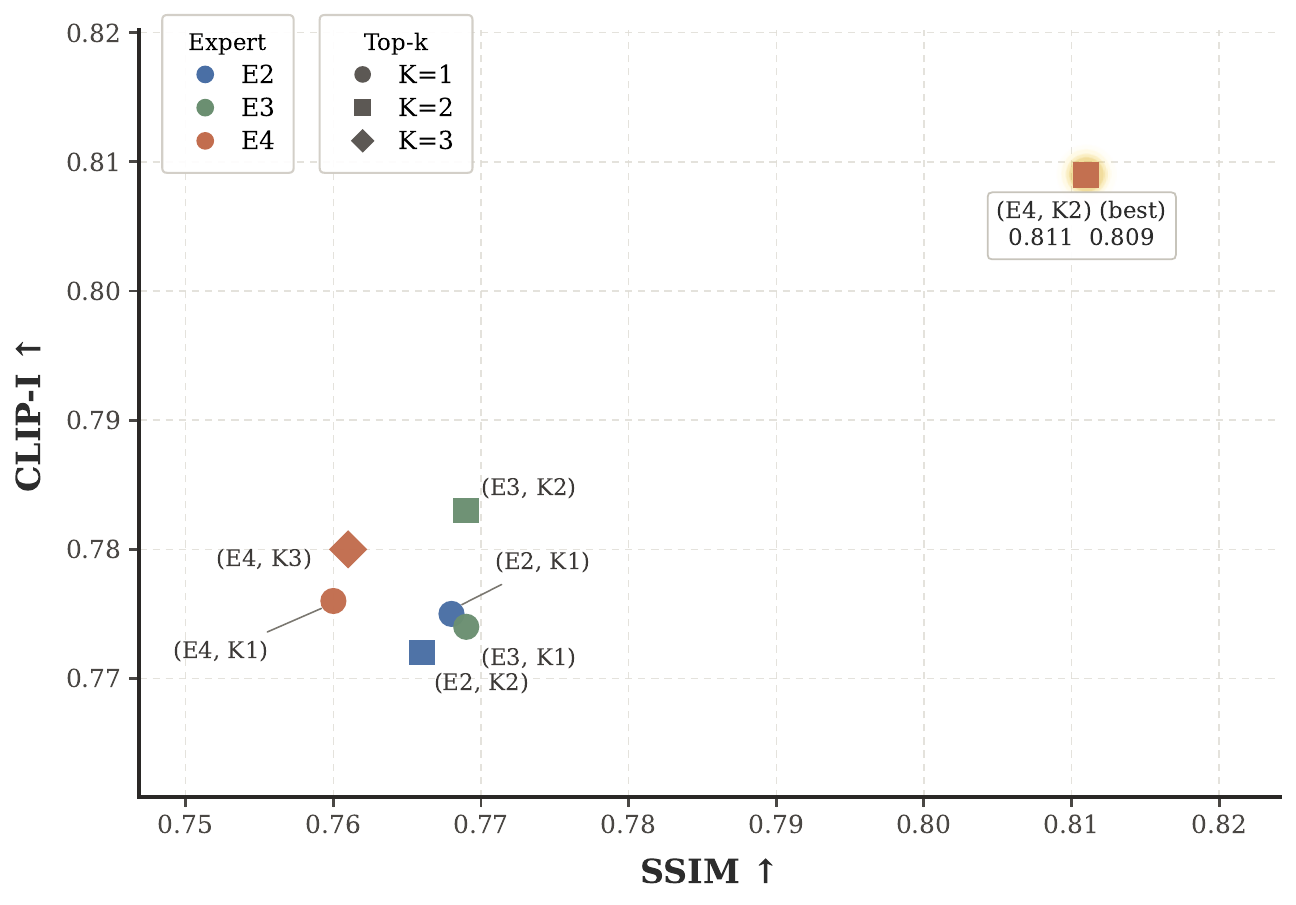}
    \caption{Ablation study on the number of experts ($n$) and the top-$k$ routing parameters.}
    \label{fig:para}
\end{figure}

\noindent \textbf{Impact of Expert Count and Top-$k$ Routing.}
To evaluate the hyper-parameters of our trait-routing attention module, we analyze the impact of the number of experts $n$ and the top-$k$ routing strategy. As shown in Figure~\ref{fig:para}, the (E4, K2) configuration achieves the best overall performance in both SSIM and CLIP-I. Specifically, a limited number of experts lacks the capacity to adequately decouple complex attributes. Although we limit the maximum number of experts to $n=4$ due to computational resource constraints, it still provides sufficient specialized pathways. Regarding the routing strategy, top-1 is overly rigid and discards complementary features, whereas top-3 introduces redundancy and semantic entanglement. The top-2 strategy strikes the optimal balance between feature diversity and precise routing. Consequently, we adopt $n=4$ and top-2 as our default setting.

\begin{figure}
    \centering
    \includegraphics[width=0.98\linewidth]{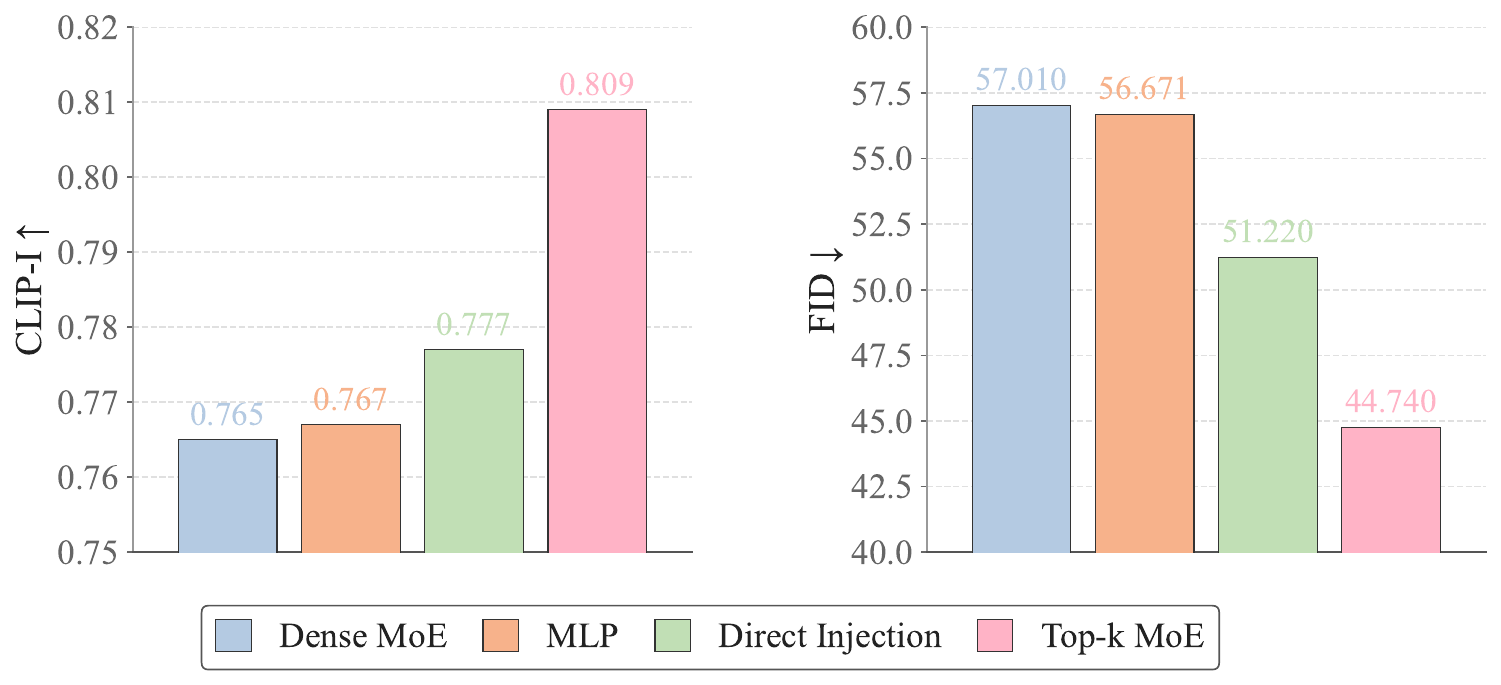}
    \caption{Analysis of CLIP-I (left) and FID (right) scores across various injection strategies.}
    \label{fig:injection_type}
\end{figure}

\noindent \textbf{Injection Strategies.}
The feature injection strategy plays a crucial role in determining the final image generation quality and the degree of feature alignment. We evaluate four feature injection strategies: dense MoE, MLP, direct injection, and our proposed Top-$k$ MoE. As illustrated in Figure ~\ref{fig:injection_type}, naive direct feature injection struggles to reconcile the inherent trade-off between semantic alignment and generative fidelity. Paradoxically, despite augmenting model capacity, both Dense MoE and conventional MLP architectures exhibit a noticeable performance degradation compared to basic direct injection, yielding inferior CLIP-I and elevated FID scores. This counterintuitive finding suggests that indiscriminate feature activation fails to assimilate external context effectively; rather, it introduces deleterious noise that compromises the target generation distribution. Conversely, our proposed Top-$k$ MoE paradigm demonstrates profound superiority by achieving the optimal CLIP-I and FID across all baselines. This confirms that the sparse Top-$k$ routing mechanism facilitates the precise assimilation of semantically pertinent features, thereby substantially elevating image-text consistency without penalizing the underlying generative fidelity.

\begin{figure*}
    \centering
    \includegraphics[width=0.98\linewidth]{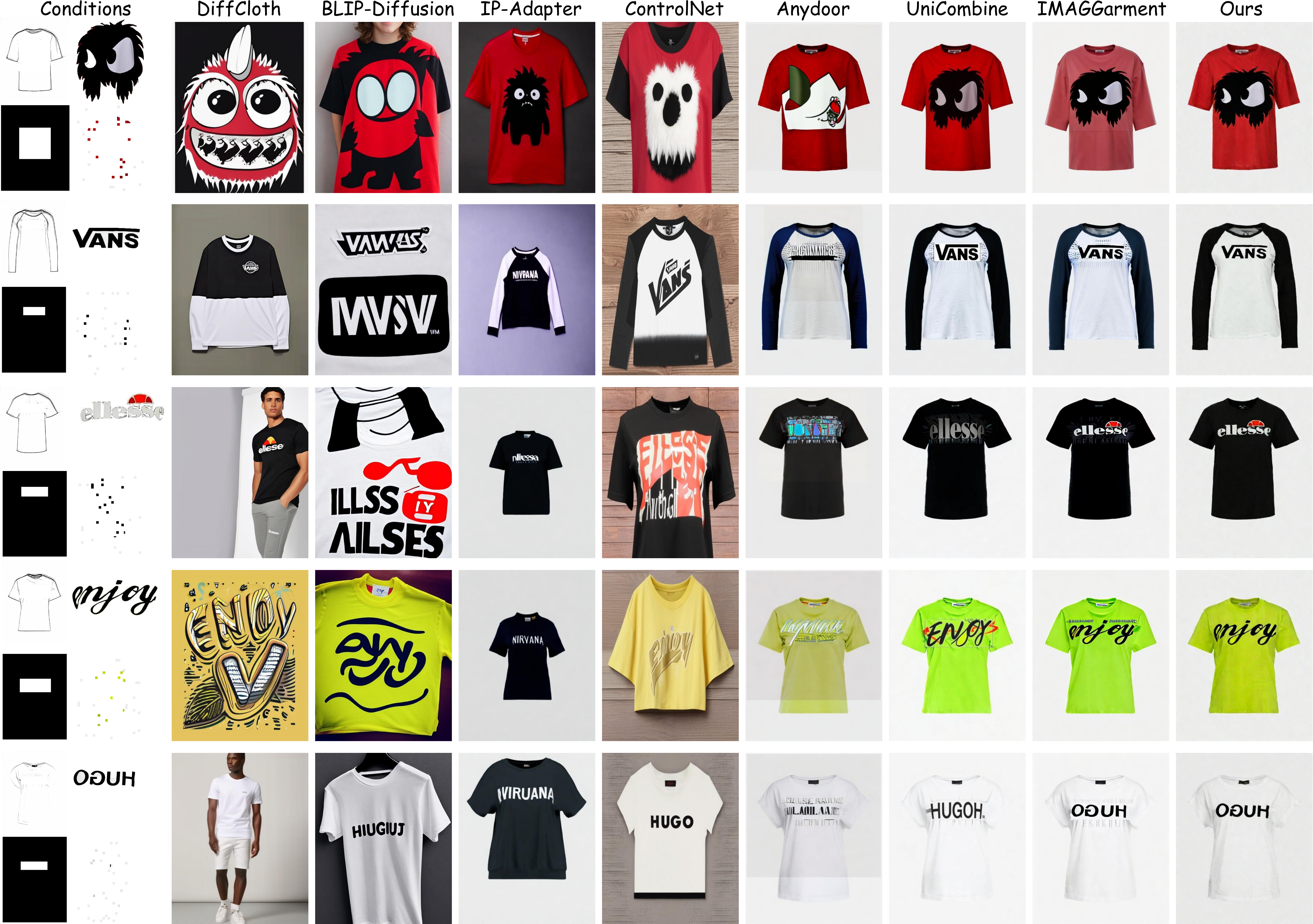}
    \caption{More qualitative comparisons I.}
    \label{fig:appendix1}
\end{figure*}

\begin{figure*}
    \centering
    \includegraphics[width=0.98\linewidth]{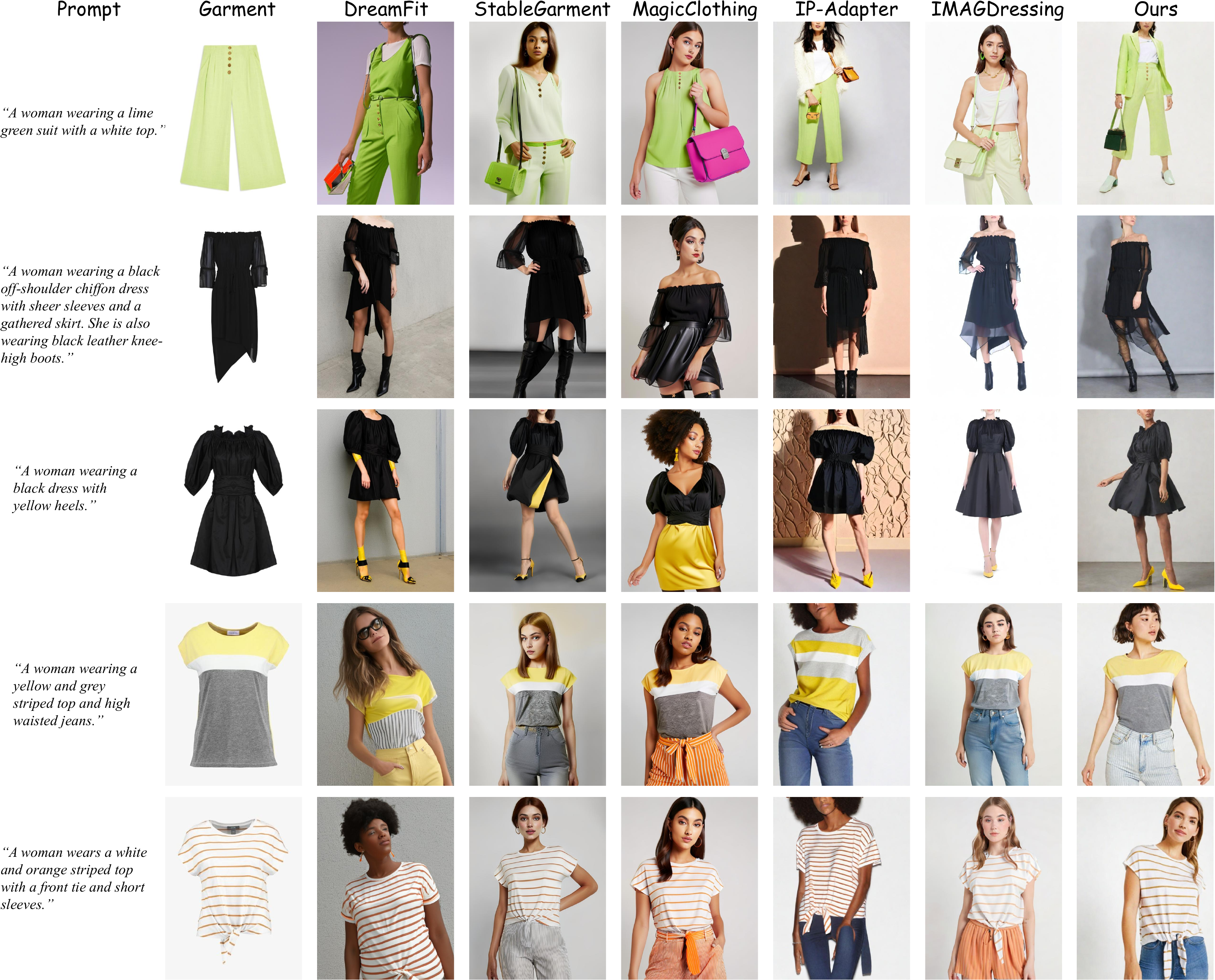}
    \caption{More qualitative comparisons II.}
    \label{fig:appendix2}
\end{figure*}

\begin{figure*}
    \centering
    \includegraphics[width=0.98\linewidth]{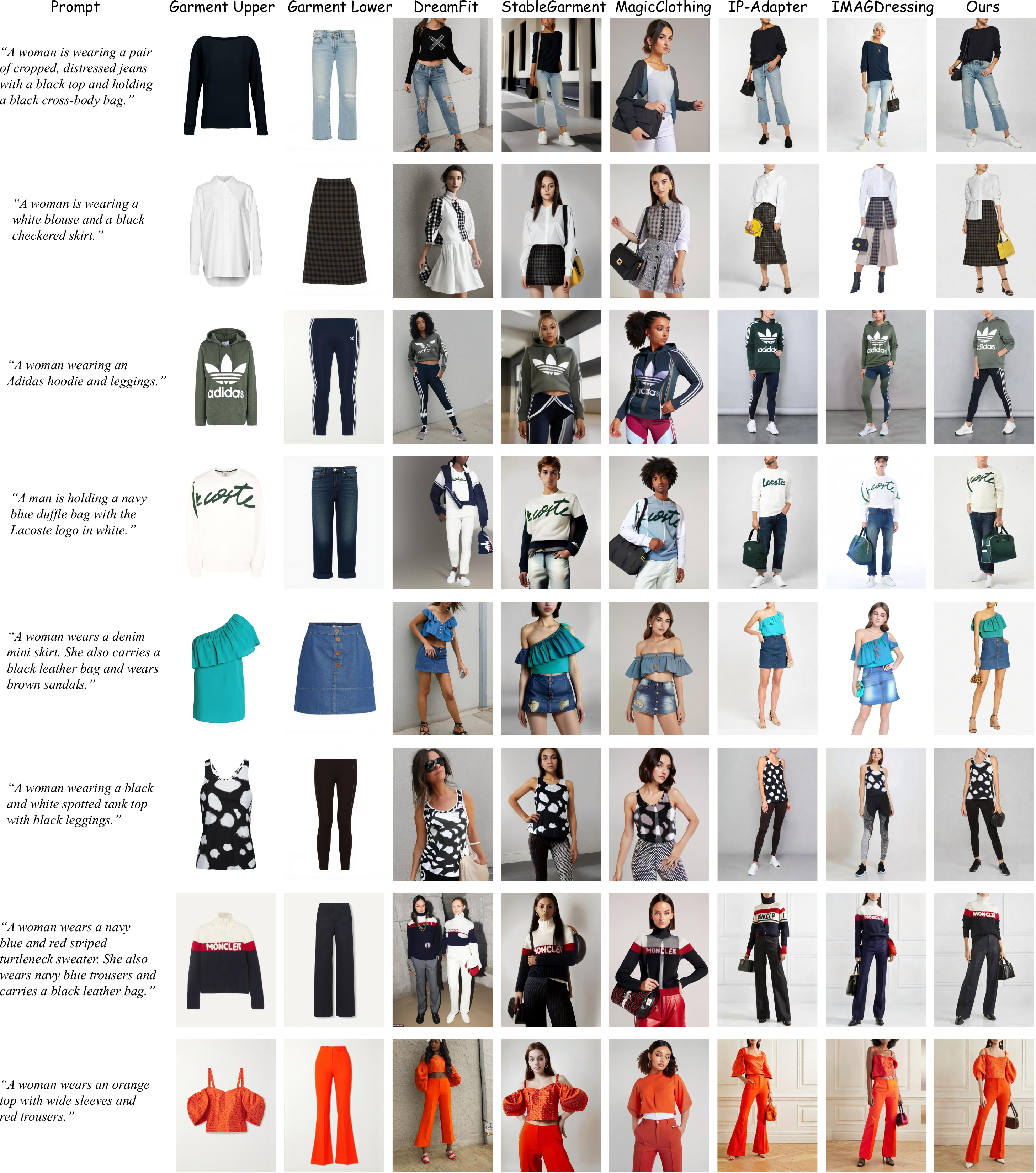}
    \caption{More qualitative comparisons III.}
    \label{fig:appendix3}
\end{figure*}

\noindent \textbf{Additional Qualitative Comparisons.}
Building upon the results discussed in the main paper, we provide further qualitative comparisons here to validate the superiority of our approach. As illustrated in Figure~\ref{fig:appendix1}, adapter-based methods for garment generation (e.g., IP-Adapter~\cite{ye2023ip} and ControlNet~\cite{zhang2023adding}) rely solely on auxiliary modules and naive addition operations for conditional feature fusion, which severely limits their ability to ensure high conditional controllability. As demonstrated by the visual comparisons for the single-garment virtual dressing task in Figure~\ref{fig:appendix2}, DreamFit~\cite{lin2025dreamfit} and StableGarment~\cite{wang2024stablegarment} fail to accurately retain the original color semantics, leading to noticeable color discrepancies and misaligned color distributions in the dressing results. As shown in Figure~\ref{fig:appendix3}, when dealing with multiple garments, MagicClothing~\cite{chen2024magic} and IMAGDressing~\cite{shen2025imagdressing} are prone to attribute confusion between different conditions, resulting in noticeable inconsistencies between the generated outputs and the reference inputs. Conversely, VersaVogue adeptly retains intricate garment details while simultaneously delivering superior global coherence and visual fidelity.


\section{More Discussion}
\label{sec:discussion}
\noindent \ding{228} \textbf{Q1: \textit{Why a unified model is required to address these two tasks?}}
The integration of garment generation and virtual dressing within a unified architectural framework stems from the need to authentically replicate the real-world fashion lifecycle, moving fluidly from conceptual design to professional presentation. By bridging these two domains, our approach enables a "design-as-you-see" experience particularly valuable in e-commerce scenarios. In such environments, users can generate personalized apparel from diverse instructions and immediately visualize the result on a target body. This holistic workflow eliminates the fragmented nature of conventional multi-stage pipelines by aligning the generative process with the natural creative sequence.

Beyond functional intuition, a unified architecture offers significant technical advantages over the practice of concatenating two independent, disparate models. A shared framework ensures inherent semantic compatibility. Whereas heterogeneous models often suffer from a representation gap that causes fine-grained textures to degrade during transfer, our VersaVogue model maintains a consistent latent space that preserves intricate garment attributes. This synergy is critical for maintaining high-fidelity details from the initial generative stage through to the final fitting.
Furthermore, this unified design effectively addresses the issue of task-specific adaptation within a consistent logic. By streamlining both tasks under the same architectural backbone, we achieve superior compatibility compared to traditional cross-framework pipelines. This ensures that the specialized features extracted for garment generation remain highly effective when repurposed for virtual dressing, significantly reducing the systemic errors that typically arise from misaligned inductive biases between different model families. Consequently, the final output is not only high-quality in isolation but also perfectly adapted to the human form, effectively balancing precise controllability with the aesthetic standards expected in professional fashion synthesis.

\noindent \ding{228} \textbf{Q2: \textit{Discussion of Concurrent Work.}}
Alongside our work, there is a notable concurrent study, Anydressing~\cite{li2025anydressing}, which focuses on multi-garment virtual dressing. Since it is a concurrent work without publicly available code or pre-trained weights, we limit our discussion to its methodology. Anydressing introduces an instance-level garment localization loss to enforce precise spatial alignment and prevent feature confusion, alongside a garment-enhanced texture learning loss to preserve high-frequency details and pattern fidelity. In contrast, our method incorporates trait-routing attention for the dynamic injection of diverse attribute features, and multi-perspective preference optimization to elevate the overall quality and coherence of the generated imagery. Diverging from Anydressing's reliance on explicit segmentation masks, our method achieves implicit feature disentanglement and precise localization, effectively bypassing the requirement for dense annotations.

\section{Theoretical Derivation of MPO}
\label{sec:dpo_loss}
Within our proposed MPO framework, we introduce a two-stage pipeline: we first automatically construct a preference dataset, and subsequently optimize the model utilizing the Direct Preference Optimization (DPO)~\cite{rafailov2023direct} loss. DPO mathematically reformulates the standard reinforcement learning from human feedback (RLHF)~\cite{ziegler2019fine} objective, eliminating the need for explicit reward modeling. In contrast to PPO-based approaches that require both a reward model and a value network, DPO directly optimizes the policy based on preference data. Formally, the standard RLHF objective seeks a policy $\pi_\theta$ that maximizes a learned reward $r_\phi(x, y)$ subject to a KL-divergence constraint against a reference model $\pi_{\text{ref}}$. This constraint prevents reward over-optimization and ensures generation diversity. The objective is defined as:
\begin{equation}
   \max_{\pi_\theta} \mathbb{E}_{x \sim \mathcal{D}, y \sim \pi_\theta(y|x)} \left[ r_\phi(x, y) \right] - 
    \beta \mathbb{D}_{\text{KL}}\left[\left(\pi_\theta(y|x) \| \pi_{\text{ref}}(y|x)\right) \right],
    \label{eq:rlhf_objective}
\end{equation}
where $\mathcal{D}$ represents the dataset of input conditions, $y$ denotes the generated output, and $\beta$ is a coefficient controlling the strength of the KL-divergence penalty $\mathbb{D}_{\text{KL}}$.

To derive the analytical solution for the optimization problem in Eq.~(\ref{eq:rlhf_objective}), we focus on the point-wise optimization for a specific input $x$. Since the objective function is concave with respect to the policy $\pi_\theta$, we can solve for its global maximum. By rearranging the terms, the functional objective for a fixed $x$ can be expressed as:

\begin{equation}
    \mathcal{J}(\pi) = \sum_y \pi(y|x) \left[ r_\phi(x, y) - \beta \log \frac{\pi(y|x)}{\pi_{\text{ref}}(y|x)} \right].
\end{equation}

To facilitate the derivation, we define the partition function $Z(x)$ as:
\begin{equation}
    Z(x) = \sum_y \pi_{\text{ref}}(y|x) \exp\left( \frac{1}{\beta} r_\phi(x, y) \right).
\end{equation}

To simplify the objective, we define a theoretical distribution $\pi^*$ based on the Gibbs formulation:
\begin{equation}
    \pi^*(y|x) = \frac{1}{Z(x)} \pi_{\text{ref}}(y|x) \exp\left( \frac{1}{\beta} r_\phi(x, y) \right).
    \label{eq:optimal_policy_def}
\end{equation}

Substituting $\pi^*$ back into the objective function, we can rewrite $\mathcal{J}(\pi)$ in terms of the Kullback-Leibler (KL) divergence between the current policy $\pi$ and the optimal policy $\pi^*$:

\begin{align}
    \mathcal{J}(\pi) &= \beta \sum_y \pi(y|x) \left[ \frac{1}{\beta}r_\phi(x, y) - \log \pi(y|x) + \log \pi_{\text{ref}}(y|x) \right] \nonumber \\
    &= \beta \sum_y \pi(y|x) \log \left( \frac{\pi_{\text{ref}}(y|x) \exp\left( \frac{r_\phi(x, y)}{\beta} \right)}{\pi(y|x)} \right) \nonumber \\
    &= \beta \sum_y \pi(y|x) \log \left( \frac{Z(x) \pi^*(y|x)}{\pi(y|x)} \right) \nonumber \\
    &= \beta \log Z(x) - \beta \mathbb{D}_{\text{KL}}\left( \pi(y|x) \| \pi^*(y|x) \right).
\end{align}

Since $\beta > 0$ and the KL divergence is non-negative ($\mathbb{D}_{\text{KL}} \geq 0$), the objective is maximized if and only if the KL divergence term is minimized to zero. Therefore, the optimal policy $\pi_{\theta}^*$ satisfies $\pi_{\theta}^* = \pi^*$. This yields the analytical solution for the optimal policy in terms of the reward function and the reference model:

\begin{equation}
    \pi_{\theta}^*(y|x) = \frac{1}{Z(x)} \pi_{\text{ref}}(y|x) \exp\left( \frac{r_\phi(x, y)}{\beta} \right).
    \label{eq:optimal_solution_explicit}
\end{equation}

This result implies that the optimal policy is an exponentially tilted version of the reference policy, scaled by the reward values. This explicit mapping between the reward function and the optimal policy is the key insight that allows DPO to eliminate the need for a separate reward model.
With the analytical solution for the optimal policy established in Eq.~(\ref{eq:optimal_solution_explicit}), we can now derive the DPO objective. The key idea is to invert the relationship to express the reward function $r_\phi(x, y)$ in terms of the optimal policy $\pi^*$ and the reference policy $\pi_{\text{ref}}$. By taking the logarithm of both sides of Eq.~(\ref{eq:optimal_solution_explicit}) and rearranging the terms, we obtain:

\begin{equation}
    r_\phi(x, y) = \beta \log \frac{\pi^*(y|x)}{\pi_{\text{ref}}(y|x)} + \beta \log Z(x).
    \label{eq:reward_reparam}
\end{equation}

This equation demonstrates that the reward for a completion $y$ given condition $x$ can be implicitly defined by the log-ratio of the optimal policy probability to the reference policy probability, offset by the partition function $Z(x)$.
Next, we consider the preference data used in RLHF. The standard reward model is typically trained using the Bradley-Terry (BT)~\cite{bradley1952rank} model, where the probability that a human prefers a winning response $y_w$ over a losing response $y_l$ is given by:

\begin{equation}
    p(y_w \succ y_l | x) = \sigma\left( r_\phi(x, y_w) - r_\phi(x, y_l) \right),
    \label{eq:bt_model}
\end{equation}

where $\sigma(z) = \frac{1}{1 + e^{-z}}$ is the logistic sigmoid function.
We now substitute the re-parameterized reward formulation from Eq.~(\ref{eq:reward_reparam}) into the BT model in Eq.~(\ref{eq:bt_model}). A crucial observation is that the partition function term $\beta \log Z(x)$ depends only on the input $x$ and is independent of the output $y$. Consequently, when calculating the difference between the rewards of the winning and losing responses, the $Z(x)$ terms cancel out:

\begin{align}
    & r_\phi(x, y_w) - r_\phi(x, y_l) \nonumber \\
    &= \bigg( \beta \log \frac{\pi^*(y_w|x)}{\pi_{\text{ref}}(y_w|x)} + \beta \log Z(x) \bigg) \nonumber \\
    &\quad - \bigg( \beta \log \frac{\pi^*(y_l|x)}{\pi_{\text{ref}}(y_l|x)} + \beta \log Z(x) \bigg) \nonumber \\
    &= \beta \log \frac{\pi^*(y_w|x)}{\pi_{\text{ref}}(y_w|x)} - \beta \log \frac{\pi^*(y_l|x)}{\pi_{\text{ref}}(y_l|x)}.
\end{align}

This cancellation eliminates the intractable partition function and the explicit reward model entirely. By replacing the theoretical optimal policy $\pi^*$ with the parameterized policy network $\pi_\theta$ that we intend to optimize, we can formulate the probability of preference directly via the policy:

\begin{equation}
\begin{split}
    p_\theta(y_w \succ y_l | x) = \sigma \Big( &\beta \log \frac{\pi_\theta(y_w|x)}{\pi_{\text{ref}}(y_w|x)} 
    - \beta \log \frac{\pi_\theta(y_l|x)}{\pi_{\text{ref}}(y_l|x)} \Big).
\end{split}
\end{equation}

Finally, similar to the reward modeling step in RLHF, DPO optimizes the policy parameters $\theta$ by minimizing the negative log-likelihood of the observed human preference data. The final DPO loss function is formulated as:

\begin{equation}
\begin{split}
    \mathcal{L}_{\text{DPO}}= -\mathbb{E}_{x, y_w, y_l} \Big[ \log \sigma \Big( &\beta \log \frac{\pi_\theta(y_w|x)}{\pi_{\text{ref}}(y_w|x)} 
    - \beta \log \frac{\pi_\theta(y_l|x)}{\pi_{\text{ref}}(y_l|x)} \Big) \Big].
\end{split}
\label{eq:dpo_loss}
\end{equation}

This objective allows optimizing the policy network directly on preference data without training a separate reward model or using PPO, significantly simplifying the RLHF pipeline while mathematically maintaining the same optimal solution.
To adapt DPO for diffusion models, we start with the standard DPO objective defined on the marginal distribution of generated samples $x_0$. Let $\mathcal{C}^\prime = \{\mathcal{C},P\}$ denote the set of all input conditions.. The objective is to maximize the likelihood of the preferred sample $x_0^w$ over the dispreferred one $x_0^l$:

\begin{equation}
\begin{aligned}
\mathcal{L}_{DPO} = -\mathbb{E}_{x_0^l,x_0^w,\mathcal{C}^\prime} \log \sigma \left(
    \beta \mathbb{E} \Big[ 
     \log \frac{p_{\theta}(x_{0}^w|\mathcal{C}^\prime)}{p_{ref}(x_{0}^w|\mathcal{C}^\prime)} - \log \frac{p_{\theta}(x_{0}^l|\mathcal{C}^\prime)}{p_{ref}(x_{0}^l|\mathcal{C}^\prime)}
    \Big]
\right).
\end{aligned}
\label{eq:dpo_marginal}
\end{equation}

Since the exact marginal likelihood $p_\theta(x_0)$ involves an intractable integral over latent variables, we expand the expectation to the full diffusion trajectories $x_{0:T}$. The objective is reformulated using the joint probability of the reverse process:

\begin{equation}
\begin{aligned}
\mathcal{L}_{DPO} = -\mathbb{E}_{x_0^l,x_0^w} \log \sigma \left(
    \beta \mathbb{E}_{x_{0:T}^w,x_{0:T}^l} \Big[ 
     \log \frac{p_{\theta}(x_{0:T}^w)}{p_{ref}(x_{0:T}^w)} - \log \frac{p_{\theta}(x_{0:T}^l)}{p_{ref}(x_{0:T}^l)}
    \Big]
\right).
\end{aligned}
\label{eq:dpo_trajectory}
\end{equation}

However, the expectation over trajectories lies inside the non-linear sigmoid function. By applying Jensen's Inequality to the convex function $f(x) = -\log \sigma(x)$, we derive a tractable upper bound. Furthermore, utilizing the Markov property of the diffusion process, we decompose the trajectory likelihood into single-step transition probabilities:

\begin{equation}
\begin{aligned}
    \mathcal{L}_{DPO} \leq -\mathbb{E}_{x_0^l,x_0^w,t,x_{t-1}^w \sim p_{\theta}(x_{t-1}^w|x_0^w),x_{t-1}^l \sim p_{\theta}(x_{t-1}^l|x_0^l) } \\
    \log \sigma \left( 
    \beta \, T \log \frac{p_{\theta}(x_{t-1}^w|x_t^w)}{p_{ref}(x_{t-1}^w|x_t^w)} -  \beta \, T \log \frac{p_{\theta}(x_{t-1}^l|x_t^l)}{p_{ref}(x_{t-1}^l|x_t^l)}
    \right).
\end{aligned}
\label{eq:dpo_bound}
\end{equation}

To obtain the final loss function, we recall that the reverse transition $p_\theta(x_{t-1}|x_t)$ in diffusion models is parameterized as a Gaussian distribution with mean $\mu_\theta(x_t, t)$ and variance $\sigma_t^2 \mathbf{I}$ ~\cite{ho2020denoising}:

\begin{equation}
    p_\theta(x_{t-1}|x_t) = \mathcal{N}(x_{t-1}; \mu_\theta(x_t, t), \sigma_t^2 \mathbf{I}) \propto \exp \left( -\frac{\|x_{t-1} - \mu_\theta(x_t, t)\|^2}{2\sigma_t^2} \right).
\end{equation}

Since the mean $\mu_\theta$ is parameterized to predict the noise $\epsilon_\theta$, maximizing this log-likelihood is equivalent to minimizing the weighted mean squared error (MSE) of the noise prediction. Specifically:

\begin{equation}
    \log p_\theta(x_{t-1}|x_t) \approx - \omega(\lambda_t) \|\epsilon - \epsilon_\theta(x_t, t)\|_2^2 + C,
\end{equation}
where $\omega(\lambda_t)$ is a weighting function derived from the noise schedule.
Substituting this relationship into Eq.~(\ref{eq:dpo_bound}), the log-probability ratio transforms into the difference of squared errors. Note that the negative sign from the Gaussian log-likelihood implies that a higher probability corresponds to a lower error. This allows us to express the final diffusion DPO objective as:

\begin{equation}
\begin{aligned}
    \mathcal{L}_{DPO} &= -\mathbb{E}_{x^l_0,x^w_0,  \mathcal{C}^\prime\sim \mathcal{D}, t, \epsilon}   \log \sigma \Big( -\beta T \omega(\lambda_t) \Big[ \\
    &\quad \left( \|\epsilon - \epsilon_\theta(x_{t}^w, \mathcal{C}^\prime, t) \|_2^2 - \|\epsilon - \epsilon_{ref}(x_{t}^w, \mathcal{C}^\prime, t) \|_2^2 \right) - \\
    &\quad \left( \|\epsilon - \epsilon_\theta(x_{t}^l, \mathcal{C}^\prime, t) \|_2^2 - \|\epsilon - \epsilon_{ref}(x_{t}^l, \mathcal{C}^\prime, t) \|_2^2 \right) \Big] \Big). 
\end{aligned}
\label{eq:dpo_final}
\end{equation}

\section{More Details}
\label{sec:more_details}
\noindent \textbf{Details of the Perceptual Quality Prompt.}
The detailed system prompts designed for perceptual quality evaluation within our Multi-Perspective Preference Optimization (MPO) framework are presented below. To elicit reliable and standardized assessments from the Vision-Language Model (VLM), we tailored specific evaluation protocols for two distinct tasks. For the Virtual Dressing task, the prompt establishes fine-grained 1-to-10 rating scales across three core dimensions: Human Realism, Clothing Fit, and Overall Aesthetic Quality. Conversely, for the Garment Generation task, the evaluation strictly focuses on apparel fidelity, assessing Material and Texture Realism, Structural Integrity and Drape, and Detail Fidelity on an identical 1-to-10 scale. Furthermore, across both tasks, we enforce a strict JSON output format, ensuring that the VLM's quantitative evaluations can be seamlessly integrated into our automated preference alignment pipeline.

\definecolor{AcademicBlue}{RGB}{30, 70, 110}
\definecolor{LightBlue}{RGB}{240, 245, 250}

\begin{tcolorbox}[
    breakable,             
    colback=LightBlue,      
    colframe=AcademicBlue,  
    title=Virtual Dressing Evaluation Prompt,
    fonttitle=\bfseries\large,
    coltitle=white,        
    enhanced,
    attach boxed title to top left={yshift=-2mm, xshift=2mm},
    boxed title style={colback=AcademicBlue},
    arc=2mm,               
    boxrule=0.8pt,
    left=4mm, right=4mm, top=5mm, bottom=4mm
]
\small \linespread{1.1}\selectfont

\textbf{System Prompt:} \\
You are an image quality evaluator specializing in AI-generated virtual dressing models. Your focus is on assessing the realism and correctness of human anatomy, the fit and coordination of garments with the body, and the overall aesthetic harmony of the image. \\

\textbf{Instructions:} \\
Given an AI-generated virtual dressing image, rate it from the following three aspects and based on the following rating criteria. Output the rating for each aspect without any reasons.
\begin{enumerate}
    \item \textbf{Human Realism:} whether the model's body features conform to normal human anatomy.  
    \item \textbf{Clothing Fit:} how well the garments fit and coordinate with the body, focusing on natural alignment and overall harmony.  
    \item \textbf{Overall aesthetic quality and realism:} evaluating whether the synthesis maintains structural coherence, realistic lighting, and an absence of perceptible artifacts.
\end{enumerate}

\textbf{Rating Criterion:} 
\begin{itemize}

\item \textit{Detailed description of each score for Human Realism rating:}
\begin{itemize}[leftmargin=1.5cm, noitemsep]
    \item[1:] Completely unrealistic human form 
    \item[2:] Highly unrealistic 
    \item[3:] Poor realism 
    \item[4:] Low realism 
    \item[5:] Acceptable but not good
    \item[6:] Decent realism 
    \item[7:] Good realism 
    \item[8:] Very good realism 
    \item[9:] Excellent realism 
    \item[10:] Photorealistic human rendering 
\end{itemize}

\item \textit{Detailed description of each score for Clothing Fit rating:}
\begin{itemize}[leftmargin=1.5cm, noitemsep]
    \item[1:] Unwearable 
    \item[2:] Very bad fit 
    \item[3:] Bad fit such as incorrect body ratio 
    \item[4:] Minor fitting issue 
    \item[5:] Acceptable but not good 
    \item[6:] Correct fit with no mistakes 
    \item[7:] Good fit with beautiful wearer presentation 
    \item[8:] Very good fit with innovations 
    \item[9:] Near perfect fit 
    \item[10:] Appears custom tailored with perfect fit 
\end{itemize}

\item \textit{Detailed description of each score for overall aesthetic quality and realism rating:}
\begin{itemize}[leftmargin=1.5cm, noitemsep]
    \item[1:] Completely disjointed composition  
    \item[2:] Highly inconsistent visual balance  
    \item[3:] Poor harmony and lack of cohesion  
    \item[4:] Low visual consistency  
    \item[5:] Acceptable but weak overall balance  
    \item[6:] Decent harmony with minor imbalance  
    \item[7:] Good visual balance and coordination  
    \item[8:] Very good overall harmony  
    \item[9:] Excellent coherence and aesthetic flow  
    \item[10:] Perfectly unified and visually harmonious composition
\end{itemize}
\end{itemize}
\textbf{Output Format:} \\
Enclose the JSON output within \texttt{<OUTPUT>} and \texttt{</OUTPUT>} tags, for example:

\begin{tcolorbox}[colback=white, colframe=gray!30, arc=1mm]
\texttt{<OUTPUT>} \\
\texttt{\{} \\
\texttt{\quad "Human Realism": 1-10,} \\
\texttt{\quad "Clothing Fit": 1-10,} \\
\texttt{\quad "Overall Aesthetic Quality and Realism": 1-10} \\
\texttt{\}} \\
\texttt{</OUTPUT>}
\end{tcolorbox}

\end{tcolorbox}

\begin{tcolorbox}[
    breakable,              
    colback=LightBlue,      
    colframe=AcademicBlue, 
    title=Garment Generation Evaluation Prompt,
    fonttitle=\bfseries\large,
    coltitle=white,       
    enhanced,
    attach boxed title to top left={yshift=-2mm, xshift=2mm},
    boxed title style={colback=AcademicBlue},
    arc=2mm,                
    boxrule=0.8pt,
    left=4mm, right=4mm, top=5mm, bottom=4mm 
]
\small \linespread{1.1}\selectfont

\textbf{System Prompt:} \\
You are an expert fashion critic and AI image quality evaluator specializing in generated apparel. Your focus is strictly on the quality, realism, and fidelity of the generated garment itself, regardless of the wearer or background. \\

\textbf{Instructions:} \\
Given an AI-generated image of a garment (either standalone or worn), rate the \textbf{garment quality} from the following three aspects based on the provided criteria. Output the rating for each aspect without any reasons.
\begin{enumerate}
    \item \textbf{Material and Texture Realism:} How realistic the fabric appears (e.g., correct distinct material properties like denim, silk, cotton, leather) and whether the texture is high-definition without artifacts.
    \item \textbf{Structural Integrity and Drape:} Whether the garment's construction follows logical physical rules (seams, edges, silhouette) and if the folds or drape behave naturally according to gravity and body shape.
    \item \textbf{Detail Fidelity:} The clarity and correctness of specific design elements such as buttons, zippers, logos, stitching, patterns, and complex accessories.
\end{enumerate}

\textbf{Rating Criterion:} 
\begin{itemize}
    
\item \textit{Detailed description of each score for Material and Texture Realism rating:}
\begin{itemize}[leftmargin=1.5cm, noitemsep]
    \item[1:] Plastic-like, blurry, or completely unrecognizable material.
    \item[2:] Severe artifacts, looks like low-res texture mapping.
    \item[3:] Poor material definition, lacks tactile visual quality.
    \item[4:] Low realism, flat lighting on fabric.
    \item[5:] Acceptable, recognizable fabric type but lacks depth.
    \item[6:] Decent realism with basic texture visibility.
    \item[7:] Good representation of fabric weight and sheen.
    \item[8:] Very good, distinct weave or leather grain visible.
    \item[9:] Excellent, highly tangible material properties.
    \item[10:] Indistinguishable from a high-end product photograph.
\end{itemize}

\item \textit{Detailed description of each score for Structural Integrity and Drape rating:}
\begin{itemize}[leftmargin=1.5cm, noitemsep]
    \item[1:] Completely broken geometry, floating parts, or impossible shapes.
    \item[2:] Major structural hallucinations (e.g., extra sleeves, melted edges).
    \item[3:] Distorted silhouette, illogical seam placement.
    \item[4:] Minor structural warping or unnatural stiffness.
    \item[5:] Acceptable shape but lacks natural gravity or flow.
    \item[6:] Correct structure with basic, logical folds.
    \item[7:] Good structural logic, natural draping around contours.
    \item[8:] Very good, complex folds and precise tailoring lines.
    \item[9:] Near perfect physical simulation of cloth dynamics.
    \item[10:] Flawless construction and physics-compliant draping.
\end{itemize}

\item \textit{Detailed description of each score for Detail Fidelity rating:}
\begin{itemize}[leftmargin=1.5cm, noitemsep]
    \item[1:] No details, smeared or garbled patterns.
    \item[2:] Very messy details (e.g., melted buttons, incoherent text).
    \item[3:] Blurry logos or jagged patterns.
    \item[4:] Low fidelity, small details are washed out.
    \item[5:] Acceptable clarity, major details are present but soft.
    \item[6:] Decent sharpness, recognizable patterns and hardware.
    \item[7:] Good definition of accessories (zippers, pockets) and prints.
    \item[8:] Very sharp, intricate patterns and stitching are visible.
    \item[9:] Excellent preservation of fine details and complex graphics.
    \item[10:] Macro-lens level clarity on all intricate design elements.
\end{itemize}
\end{itemize}

\textbf{Output Format:} \\
Enclose the JSON output within \texttt{<OUTPUT>} and \texttt{</OUTPUT>} tags, for example:

\begin{tcolorbox}[colback=white, colframe=gray!30, arc=1mm]
\texttt{<OUTPUT>} \\
\texttt{\{} \\
\texttt{\quad "Material and Texture Realism": 1-10,} \\
\texttt{\quad "Structural Integrity and Drape": 1-10,} \\
\texttt{\quad "Detail Fidelity": 1-10} \\
\texttt{\}} \\
\texttt{</OUTPUT>}
\end{tcolorbox}

\end{tcolorbox}
\noindent \textbf{Details about the Preference Dataset.}
\begin{figure}
    \centering
    \includegraphics[width=0.98\linewidth]{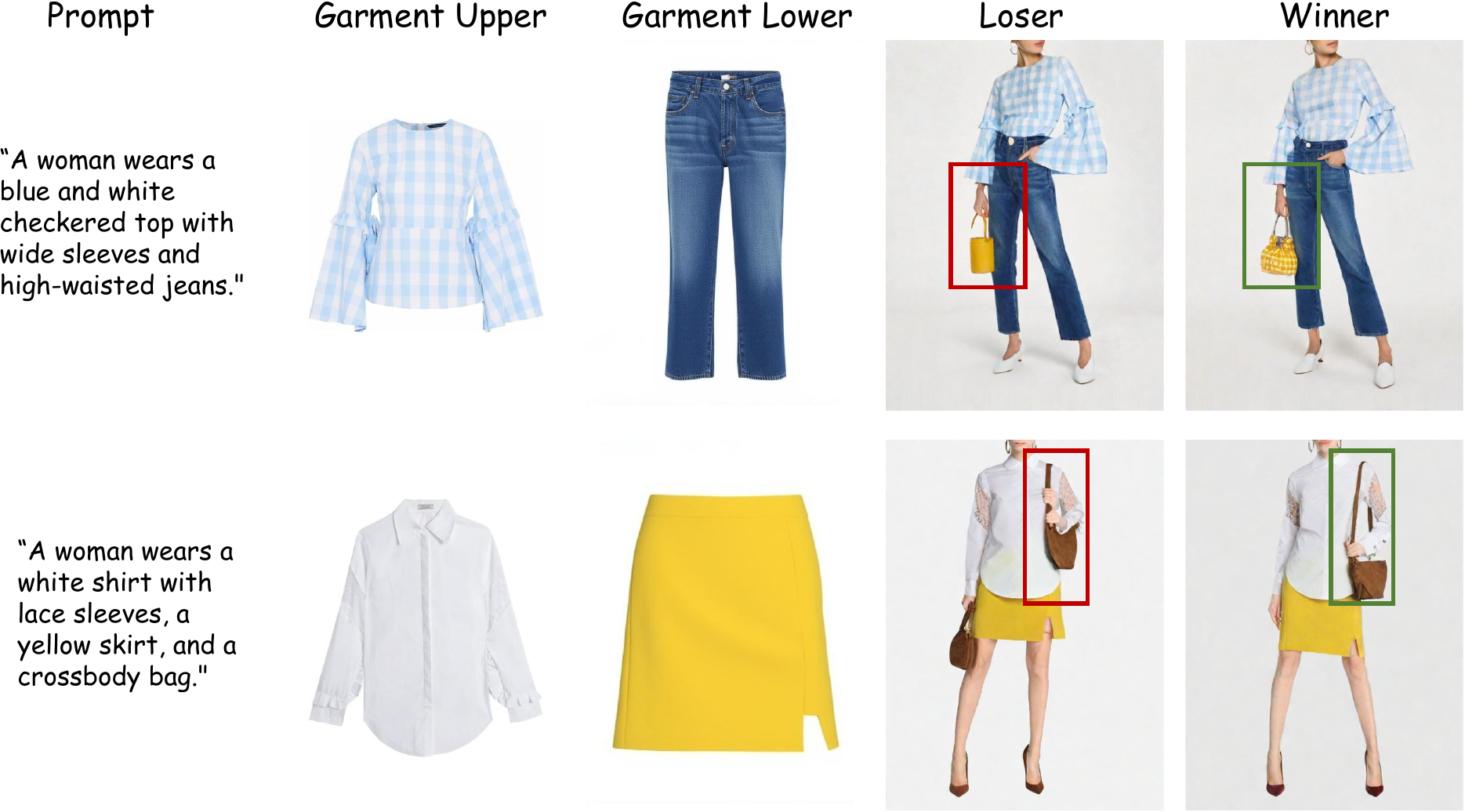}
    \caption{Examples from the preference dataset synthesized via the MPO strategy.}
    \label{fig:preference_dataset}
\end{figure}
To synthesize visually harmonious images that better align with human aesthetics, we leverage MPO to automatically curate a preference dataset, which is coupled with the DPO objective for model optimization. As illustrated in Figure~\ref{fig:preference_dataset}, the dataset comprises paired generation results categorized into "Winner" (good) and "Loser" (bad) samples. Notably, these image pairs are generated under identical conditions, including the same text prompts and reference garments, but with different random seeds. The selection criteria penalize structural distortions and local artifacts to ensure high perceptual quality. For instance, in the highlighted regions of Figure~\ref{fig:preference_dataset}, the "Loser" sample exhibits severe structural artifacts and unnatural physical interactions, such as the distorted hand grasping the accessory. Conversely, the "Winner" sample demonstrates superior visual harmony, coherent anatomical structures, and seamless integration of the reference garments. By capturing these granular differences in generation quality, the curated dataset provides a robust, human-aligned learning signal for the subsequent fine-tuning stage.

\noindent \textbf{Definition of Routing Entropy.}
To quantitatively analyze the token allocation behavior within our trait-routing attention module, we compute the routing entropy, where a higher value indicates a more uniform distribution of tokens across all branches, and a lower value signifies a concentrated, highly specialized routing behavior.
Let $l \in \{1, 2, \dots, L\}$ denote the index of the network layer. For a given garment category, let $T$ be the total number of valid visual tokens across all input images. In our routing mechanism, each token $t$ at layer $l$ is discretely assigned to its top-$k$ experts among the $n$ available branches (in our implementation, $n=4$). 
To formalize this, we define an indicator function $\mathbb{I}_{l}(t, i)$, which equals $1$ if the $i$-th expert is selected within the top-$k$ experts for token $t$ at layer $l$, and $0$ otherwise. First, we calculate the normalized routing frequency $p_{l,i}$ for the $i$-th expert at layer $l$. Since each of the $T$ tokens selects $k$ experts, the total number of expert assignments is $k \times T$. The frequency distribution $p_{l,i}$ is computed as:
\begin{equation}
    p_{l,i} = \frac{1}{k \cdot T} \sum_{t=1}^{T} \mathbb{I}_{l}(t, i)
\end{equation}
where $p_{l,i}$ represents the empirical probability that an assigned token is routed to the $i$-th expert at layer $l$. This guarantees that $\sum_{i=1}^n p_{l,i} = 1$.
The routing entropy $H_l$ for layer $l$ is defined as:
\begin{equation}
    H_l = -\sum_{i=1}^{n} p_{l,i} \ln(p_{l,i})
\end{equation}

\noindent \textbf{Metrics.}
We employ representative metrics to assess both the quality of synthesized images and conditional alignment. Fréchet Inception Distance (FID)~\cite{heusel2017gans} and Learned Perceptual Image Patch Similarity (LPIPS)~\cite{zhang2018unreasonable} are utilized across both the garment generation and virtual dressing tasks to gauge visual realism and perceptual fidelity. Additionally, we employ distinct metrics for different tasks: for garment generation, Color Structure Similarity (CSS)~\cite{zeng2014color} and Logo Location Accuracy (LLA)~\cite{fujitake2024rl} are utilized to verify chromatic alignment and spatial precision, respectively. Meanwhile, the virtual dressing task incorporates Structural Similarity Index Measure (SSIM)~\cite{wang2004image} to evaluate structural preservation and CLIP-Image Similarity (CLIP-I)~\cite{radford2021learning} to ensure the high-level semantic identity of the reference garment is maintained.

\section{Limitations and Future Work}
\label{sec:limitation}
\begin{table}
    \centering
    \caption{Computational efficiency of our method. The inference time is measured on a single NVIDIA A800 GPU.}
    \label{tab:efficiency_comparison}
    \begin{tabular}{lcc}
        \toprule
        \textbf{Method} & \textbf{Trainable Params (M)}  & \textbf{Inference Time (s)}  \\
        \midrule
        Ours       & 3644.21 & 27.740 \\
        \bottomrule
    \end{tabular}
\end{table}
Despite the superior performance and high-fidelity generation demonstrated by VersaVogue, we acknowledge several limitations in our current implementation. One significant challenge is the substantial memory overhead incurred by our framework, as detailed in Table~\ref{tab:efficiency_comparison}. The architectural design, specifically the parallel deployment of multiple UNets coupled with the Mixture-of-Experts (MoE) architecture, inevitably increases the video RAM footprint and imposes strict hardware requirements for training. However, although the total parameter count is substantial, the MoE mechanism ensures that only a subset of parameters (\emph{i.e.}, the top-2 experts) is activated during a single forward pass. Consequently, the actual computational FLOPs do not increase linearly with the parameter size. Furthermore, our current inference latency of 27.7 seconds is measured using a standard 50-step DDIM~\cite{song2022denoisingdiffusionimplicitmodels} sampler. In production environments, integrating acceleration techniques such as Latent Consistency Models~\cite{luo2023latent} (LCMs) or leveraging high-performance computing clusters for optimized deployment can significantly reduce this latency. To fundamentally address the resource bottleneck, future work will aim to design a more lightweight MoE architecture to substantially reduce the overall parameter count while maintaining high-quality generation performance.

Furthermore, while our unified framework excels in 2D garment generation and virtual dressing, its capability to model complex 3D physical properties, such as multi-view consistency and realistic spatial geometry, remains limited. The 2D-centric formulation intrinsically restricts its performance in scenarios requiring immersive 3D interactions. To address this, our future work will explore extending the framework into the 3D domain, aiming to achieve high-fidelity 3D garment synthesis and spatial virtual dressing.

\end{document}